\documentclass[letterpaper]{new_tlp}
\usepackage{times}
\usepackage{helvet}
\usepackage{courier}
\usepackage[]{natbib}
\usepackage{latexsym}
\usepackage{soul}
\usepackage{amssymb} 
\usepackage[toc,page]{appendix}
\usepackage{graphicx}
\newcommand{\Begproof}{\begin{proof}}
\newcommand{\Endproof}{\end{proof}}

\usepackage{enumitem} 

\usepackage{url}

\usepackage{todonotes}

\usepackage{color}
\usepackage[normalem]{ulem}
\usepackage{wrapfig}
\usepackage{amsmath} 

\usepackage{acronym}

\acrodef{KRR}[KR\&R]{Knowledge Representation and Reasoning}
\acrodef{RAC}[RAC]{Reasoning about Actions and Change}
\acrodef{ASP}[ASP]{Answer Set Programming}
\acrodef{CP}[CP]{Constraint Programming}
\acrodef{AI}[AI]{Artificial Intelligence}

\acrodef{IR}[IR]{Information Retrieval}
\acrodef{NLP}[NLP]{Natural Language Processing}
\acrodef{IRAC}[iRAC]{Information Retrieval with Actions and Change}
\acrodef{ACIR}[ACIR]{Actions-Centered Information Retrieval}
\acrodef{QA}[QA]{Question Answering}

\newtheorem{proposition}{Proposition}
\newtheorem{theorem}{Theorem}
\newtheorem{lemma}[theorem]{Lemma}
\newtheorem{definition}{Definition}
\newtheorem{example}{Example}
\newtheorem{corollary}{Corollary}

\newcommand{\tbeg}{\langle}
\newcommand{\tend}{\rangle}
\newcommand{\entails}{\models}

\newcommand{\lpnot}{\mbox{not}\;\,}

\newcommand{\hif}{\leftarrow}

\newcommand{\causes}{\mbox{ causes }}

\newcommand{\lif}{\mbox{ if }}
\newcommand{\imp}{\mbox{ impossible\_if }}

\newcommand{\cross}{\Join}

\def\useLongVersion{\let\ifLongVersion=\iftrue\let\ifShortVersion=\iffalse}
\def\useShortVersion{\let\ifLongVersion=\iffalse\let\ifShortVersion=\iftrue}

\useShortVersion

\newcommand{\extlit}{\lambda}

\newcommand{\query}{\boldsymbol{q}}

\begin{document}

\title{Action-Centered Information Retrieval}
  \author[M. Balduccini and E. LeBlanc]
         {MARCELLO BALDUCCINI\\
         Saint Joseph's University\\
         marcello.balduccini@gmail.com
         \and
         EMILY C. LEBLANC\\
         Drexel University\\
         ecl38@drexel.edu}



\maketitle

%
%
\begin{abstract}
Information Retrieval (IR) aims at retrieving documents that are most relevant to a query provided by a user. Traditional techniques rely mostly on syntactic methods. In some cases, however, links at a deeper semantic level must be considered. In this paper, we explore a type of IR task in which documents describe sequences of events, and queries are about the state of the world after such events. In this context, successfully matching documents and query requires considering the events' possibly implicit, uncertain effects and side-effects. We begin by analyzing the problem, then propose an action language based formalization, and finally automate the corresponding IR task using Answer Set Programming.
\\\emph{Under consideration in Theory and Practice of Logic Programming (TPLP).}

\end{abstract}


%
%
\section{Introduction}\label{sec:intro}
\ac{IR} \citep{kor97} aims at identifying, among a set of available information sources, those that are most relevant to a query provided by the user. \ac{IR} is arguably a staple of every day life -- we consult Wikipedia for general reference, doctors search private databases for patient information, and researchers use public databases to find scientific publications. \ac{IR} is also at the core of numerous commercial activities such as searching for news about business partners or competitors. 

Most \ac{IR} systems base the relevance of a source on a syntactic measurement of the overlap of terms between query and source \citep{MDRS2008}. Even advanced techniques still focus on syntactic matching, and include temporal IR \citep{C2015}, query expansion \citep{CR2012}, and graph-based term weighting \citep{RC2012}. 

Recent research \citep{gs14} has demonstrated that traditional \ac{IR} yields low accuracy when applied to documents centered on events, such as police reports, medical records, and breaking news. As one can imagine, documents centered on events occur in large quantities and often contain very valuable information. 
Consider an example related to healthcare: a radiologist might be looking for information on whether a patient was ever bedridden. This type of information is rarely stated explicitly in patient documents; rather, the radiologist is more likely to have access to documents reporting, for instance, that the patient suffered a multiple fracture at his left leg.  Such a document is indeed relevant to the radiologist's request, since the patient was likely bedridden as a result of the injury. However, determining its relevance requires linking the event of suffering a leg injury to the resulting state of the patient, which is beyond the reach of \ac{IR} techniques based on syntactic measurements. 
Similar considerations can be made  in the context of cybersecurity/cyberanalytics. Consider the case of a user asking whether a computer was left without network connectivity during a certain time frame.
A system log stating that the router, to which the computer is connected, was restarted during that period of time, is indeed relevant to the user's query. However, identifying its relevance requires the ability to consider that  a restart causes the router to transition to a state in which connected devices are without network connectivity. Again, this capability is clearly beyond the reach of traditional \ac{IR}. In fact, in practice this use case may be even more complicated if the computer is connected to the router through other devices, in which case identifying a match requires  reasoning about how the loss of connectivity propagates recursively to any device connected to the router.

\cite{gs14} proposed a new approach, called event-centered \ac{IR}, which succeeded in increasing match accuracy by means of some level of semantic analysis. However, their approach was limited to matching events mentioned in both queries and sources. This is insufficient to handle the above examples, where one needs to link events and information about the \emph{state of the world} before, during, and after the events.


In this paper, we advance this line of research by considering the case in which the goal is to match sources containing sequences of events with queries that are about the state of the world after those events. 
The examples provided above fall in this category, as well as, generally speaking, all cases in which the sources describe the history of a domain (e.g., historical documents about property sales, police reports, computer event logs) and a user is interested in those sources from which the state of the world at a certain point in time can be reconstructed (e.g.,  ``was the firewall active when the attack happened?''). 

Our approach aims to identify reasonable matches even when a definitive answer cannot be immediately found in the sources, events have complex/hidden effects, and information is incomplete. We call the corresponding reasoning task \emph{Action-Centered \ac{IR}}. 

The aim of the present paper is to provide an accurate definition of the problem, of the corresponding reasoning tasks, and of algorithms for automating the process. We begin the paper by analyzing the problem and, appealing to commonsense and intuition, identify reasonable outcomes of the task as a human reader might carry it out. For illustration purposes, we start with a simple problem, which we progressively elaborate. An exhaustive evaluation of our approach over realistic examples is beyond the scope of the paper: as discussed by \cite{gs14}, the existing benchmarks for \ac{IR} tasks are not suitable for the evaluation of semantic-level matching approaches and the development of suitable datasets is a research project in its own merit, which we will tackle and document separately. However, we conducted a preliminary investigation of the scalability of our approach, whose results are discussed later in the paper.

It should be noted that, throughout the paper, we assume that passages in natural language have already been translated into a suitable logic form, as the natural language task is orthogonal to the problem addressed here. Specifically, we assume that query and sources have already been translated to a temporally-tagged logical representation, e.g., by using techniques from \citep{nmb15,lb16,lig17}. We also assume the availability of suitable knowledge repositories 
\citep{M2006,S2008,inc16}. 
Additionally, while our work is somewhat related to research on temporal relations (e.g., Allen's interval calculus), the two differ because we focus on reasoning about events and their effects, rather than relations between events.

The main contributions of this paper are (a) the exploration of a novel, non-trivial variant of \ac{IR} in which sources include sequences of events, and queries are about the state of the world after such events; (b) the extension of techniques for representing dynamic domains to increase the flexibility of reasoning in the presence of uncertainty; (c) a formalization of the \ac{IR} task based on action languages; (d) an automated \ac{IR} procedure based on Answer Set Programming (ASP).

In the next section, we cover the analysis of the problem. Next, we present a formalization of the relevant knowledge and of the reasoning tasks underlying Action-Centered \ac{IR}. In the following section, we illustrate an ASP-based procedure for automating the reasoning processes, and demonstrate it on the toy scenarios. Then, we present an experimental evaluation of our approach. Finally,  we discuss related work and draw conclusions.

\section{Problem Analysis}\label{sec:analysis}
%

In this section, we proceed by example to an analysis of the problem of Action-Centered \ac{IR} and discuss, in commonsensical terms, the underlying reasoning tasks. Let us start with the following:
\begin{example}\label{ex:date}
{\em The user's query, $\query$, is ``Is John married?'' Available information sources are:\\
\hspace*{.1in}$\mathcal{S}_1$: ``John went on his first date with Mary.''\\
\hspace*{.1in}$\mathcal{S}_2$: ``John read a book.''\\
We want to determine which source is most relevant to $\query$.
}
\end{example}
The query refers to the current state of the world, which with some approximation we can identify with the final state of the world in the sources. The sources describe events that occurred over time.  Neither source mentions being married, making syntactic-based methods unfit for the task. However, from an intuitive perspective, $\mathcal{S}_1$ is more relevant to $\query$ than $\mathcal{S}_2$. In fact, $\mathcal{S}_1$, together with commonsense knowledge that married people typically do not go on first dates, provides a strong indication  that John is not married. $\mathcal{S}_2$, on the other hand, provides no relevant information.

In this simple example, one can not only identify $\mathcal{S}_1$ as the most relevant source, but also look for an accurate answer to the question. The simplicity of the example blurs the line between \ac{IR} and question answering. In general, however, providing an accurate answer requires a substantial amount of reasoning to be carried out once a relevant source has been identified, as well as deep understanding of the content of the source and a large amount of world knowledge -- something that is still challenging for state-of-the-art approaches. Thus, in this paper, \emph{we assume that a reader with human-level intelligence will later find accurate answers by studying the sources identified as relevant by our approach. We focus on defining techniques that provide the reader with a ranking of the sources based on our expectation that answers may be found in them.}

The previous example allows us to establish a first, high-level characterization of the task we aim to study, as one in which we are given a query $\query$ and a collection of sources $\mathcal{S}_1, \ldots, \mathcal{S}_n$, and are asked to produce scores $s_1, \ldots, s_n$ ranking each source based on its relevance to the problem of finding an answer to $\query$. If we adopt the convention that $0$ is the best possible score and $\infty$ the worst, then it is conceivable that, in Example \ref{ex:date}, $\mathcal{S}_1$ should be assigned a score of $0$ and $\mathcal{S}_2$ a score of $\infty$ to indicate complete irrelevance.

As in traditional \ac{IR}, the sources are ranked based on their respective score. We expect that, in the long-term, both syntactic and semantic aspects will have to be taken into account to determine scores for the documents. Thus, below, we use the term ``semantic score'' when we refer to the score assigned to documents by the techniques we study.
It is worth stressing the difference between the task at hand and question answering, where the goal is to produce a definitive answer. At the end of the process we consider here, the answer to $\query$ may still be unknown, but there will be reason to believe that careful study of the sources identified as relevant will lead to an answer.

Next, we consider a number of examples and corresponding expectations. Based on the examples, later we propose a formalization of the reasoning processes.
Example \ref{ex:date} showed that the event of going on a first date may lead us to infer that John is not married. But how can one reach such conclusion? One option is to reason by cases, and consider two possible views of the world: one in which John is married at the beginning of the story, and one in which he is not.  Commonsense tells us that the action\footnote{From now on, we will use action and event as synonyms.} of going on a first date is not executable when married. Hence, the view in which John is initially married is inconsistent with the source. So, we conclude that John must not have been married in the initial state. Given further knowledge that one does not get married on a first date, one can infer that John remains not married after the date. Thus, the source provides evidence that a reader can use to answer the query.

From a technical perspective, the example highlights the importance of being able to reason  by cases, to reason about the executability of actions, and to propagate the truth of properties of interest over the duration of the story. Note, however, that reasoning by cases is sometimes misleading. Consider $\mathcal{S}_2$ from Example \ref{ex:date}: reasoning by cases leads to the same two possible initial states. Since reading does not affect married status, there are two ending states for the story. This might be taken as an indication that the source provides \emph{some} useful evidence for a reader, but it is clear intuitively that $\mathcal{S}_2$ is, in fact, irrelevant. Next, let us consider if, and how, the previous query should match a more complex document. For the sake of this example, let us assume the existence of a fictitious country C, whose laws allow plural marriage.
\begin{example}\label{ex:date-mormon}
{\em
$\query$: Is John married?
\\
$\mathcal{S}$: John, who lives in country C, just went on his first date with Mary.
}
\end{example}
In this case, $\mathcal{S}$ does not provide useful information towards answering $\query$. John is from C, where plural marriage is allowed, and knowledge about plural marriage yields that being married does not preclude a married person from going on a first date. The example also demonstrates the importance of reasoning about \emph{default statements} (statements that are \emph{normally} true) and their exceptions. The fact that married people typically do not go on first dates is an instance of a default statement, and an inhabitant of C constitutes an exception to it. Similarly to $\mathcal{S}_2$ from the previous example, reasoning by cases may be somewhat misleading, as it may suggest that the source provides some evidence useful to answering the question. Rather than reasoning by cases, it appears to be more appropriate to state that whether John is initially married is \emph{unknown}. The lack of knowledge is propagated to the final state, given that going on a date has no effect on it in the present context. The source is thus irrelevant and should receive a semantic score of $\infty$. Note the striking difference in scores between $\mathcal{S}_1$ from the previous example and the current source: it appears that in some cases reasoning by cases is useful, while in others reasoning explicitly about lack of knowledge is more appropriate. In the next section, we provide a characterization of reasoning matching this intuition. Next, we investigate the role of the effects of actions.

\begin{example}\label{ex:married-mormon}
{\em
$\query$: Is John married?
\\
$\mathcal{S}$: John, who lives in country C, recently went on his first date with Mary. A week later, they tied the knot in Las Vegas.
}
\end{example}
Obviously, a first indication of relevance can be obtained with shallow reasoning and syntactic matching: ``tying the knot'' is a synonym of ``getting married,'' and ``getting married'' and ``being married'' share enough similarities to make a match likely. However, we are interested in more sophisticated reasoning. In the initial state, John may or may not be married due to his country's laws. Similarly to Example \ref{ex:date}, John's married status persists in the state following the first date. Tying the knot, however, has the effect of making John married in the resulting state. Hence, $\mathcal{S}$ is indeed relevant to $\query$. Intuitively, its semantic score should be equal to that of $\mathcal{S}_1$ from Example \ref{ex:date}. This demonstrates the importance of keeping track of the changes in the truth of the relevant properties over time. The next example takes this argument one step further.
\begin{example}\label{ex:filed}
{\em
$\query$: Is John married?
\\
$\mathcal{S}$: John recently went on his first date with Mary. A week later, they tied the knot in Las Vegas. A month later, they filed for divorce.
}
\end{example}
Here, we assume that filing for divorce does not immediately cause the spouses to be divorced. For simplicity, we adopt a view in which filing has a non-deterministic effect: in the resulting state, it is equally likely for the spouses to be married or not. The relevance of $\mathcal{S}$ to $\query$ is not as straightforward as in previous cases. It is true that, at the end of the story, it is unknown whether John is married. On the other hand, the story still provides some information pertaining to John's married status -- certainly, more than source $\mathcal{S}_2$ (``John read a book'') from Example \ref{ex:date} or the source from Example \ref{ex:date-mormon} (``John, who lives in country C, just went on his first date with Mary.'').

One way to make a distinction between the two cases is to consider that, if $\mathcal{S}$ from Example \ref{ex:filed} is provided to a reader, and the reader manages to determine if the filing action succeeded (e.g., by gathering additional evidence), $\mathcal{S}$ will immediately allow the reader to answer $\query$. Differently from the previous examples, knowing that filing occurred is \emph{essential} to allowing a reader to answer the question. 
Hence, while $\mathcal{S}$ is not as relevant to $\query$ as other sources we have considered, it is still somewhat relevant.
\ifLongVersion

This example highlights the fact that it is indeed reasonable to consider possible outcomes of non-deterministic actions, and to reason by cases in doing so. If the source does not provide sufficient information to decide between the outcomes, its relevance should be reduced, as the reader will need to conduct further investigations.

\fi
\ifLongVersion
Along the same lines, it seems that there is utility in making assumptions in order to find a match between source and query. Consider the following example:
\begin{example}\label{ex:martian}
\ \\
$\query$: Is John married?\\
$\mathcal{S}$: John brought Mary flowers.
\end{example}
\textcolor{red}{DROP THIS EXAMPLE IF WE DROP THE IDEA OF FORCING DEFAULT FLUENTS.} The source is seemingly unrelated to the query. However, suppose that world knowledge includes information that, according to the costume of country X, bringing someone flowers causes the two to immediately get married. Also, suppose that the majority of people are not from country X, i.e., our world knowledge includes a default ``normally, people are not from country X.'' 

If a reader with human-level intelligence were able to determine, later, that John is from country X, that would immediately lead to concluding that John is married. Hence, $\mathcal{S}$ is partially related to $\query$.

In  conclusion, in attempting to identify a match between $\mathcal{S}$ and $\query$, it seems acceptable to hypothesize that the source is about an atypical situation. As in the previous example, when doing so, the relevance of the source should be reduced accordingly.

\fi
This will have to be reflected in the score assigned to the source, which should be higher than the $0$ assigned to $\mathcal{S}_1$, but obviously smaller than $\infty$ because the source is indeed relevant. Next, we propose a formalization that captures the behaviors described.

\section{Formalization of the Action-Centered \ac{IR}}\label{sec:formalization}
One may note that carrying out the reasoning discussed above requires considering how actions may affect the state of the world in possibly indirect and intricate ways. For this reason, our formalization of the reasoning task at the core of Action-Centered \ac{IR} leverages techniques from the research on reasoning about actions and change, and specifically action language $\mathcal{AL}$ \citep{bg00},  approximated representations \citep{mts07} and evidence-based reasoning \citep{gb02}. These techniques rely on a graph-based representation of the evolution of the state of the world over time in response to the occurrence  of actions. We adopt and expand this approach. Specifically, similarly to the approach by \cite{mts07}, our formalization enables reasoning explicitly about lack of knowledge. Differently from it, however, we allow a reasoner to reason by cases whenever needed. This is applied to knowledge about both initial state and effects of actions. Our approach also leverages evidence-based reasoning to rule out some of the cases considered. Finally, we adopt $\mathcal{AL}$ as the underlying formalism, but expand it for an explicit characterization of  non-deterministic effects and we allow hypothesizing about exceptional/atypical circumstances, eventually linking them to the relevance of sources. Differently from  $\mathcal{AL}$, our language is defined so that, in the presence of actions with non-deterministic effects, it is possible to reason by cases as well as by explicitly characterizing lack of knowledge. The syntax of the language,  which we call $\mathcal{AL}_{I\!R}$, is described next, followed by its semantics.

Let $\mathcal{F}$ be a set of symbols for \emph{fluents} and $\mathcal{E}$ be a set of  symbols for \emph{elementary actions}.
Fluents are boolean properties of the domain, whose truth value may change over time. An \emph{action} is a set of elementary actions. With slight abuse of notation, we denote a singleton action by its  element. 

A \emph{fluent literal} is a fluent $f$ or its negation $\neg f$. The complement of $f$ (written $\overline{f}$) is $\neg f$, and vice-versa. The set of literals formed from $\mathcal{F}$ is denoted by $Lit$. An \emph{extended (fluent) literal} is either a fluent literal or an expression of the form $u(f)$, intuitively meaning that it is unknown whether fluent $f$ is true or false. An expression  $u(f)$ is called a \emph{proper extended literal}. 
%

A \emph{signature} is a tuple $\Psi=\tbeg \mathcal{F}, \mathcal{} \mathcal{E} \tend$, whose components are defined above. Given a signature, the \emph{laws} of $\mathcal{AL}_{I\!R}$ are statements of the form:
\begin{align}
\label{eq:dynamic-law}
& e \causes \extlit \lif l_1, l_2, \ldots, l_n \\
\label{eq:state-constraint}
& l_0 \lif l_1, \ldots, l_n \\
\label{eq:exec-condition}
& e \imp l_1,  \ldots, l_n
\end{align}
where $\extlit$ is an extended literal, $l_1, \ldots, l_n$ are fluent literals, and $e$ is an elementary action\footnote{We focus on elementary actions for simplicity of presentation. Expanding the laws to allow non-elementary actions is not difficult.}. Law (\ref{eq:dynamic-law}) is called \emph{dynamic (causal) law}. If $\extlit$ is a fluent literal, the law intuitively says that, if action $e$ is executed in a state in which $l_1, \ldots, l_n$ hold, then $\extlit$ will hold in the next state. If $\extlit$ is a proper extended literal $u(f)$, the law intuitively states that the action affects the truth of $f$ non-deterministically. $\extlit$ is called the \emph{consequence of the law}. The action of filing for divorce from Example \ref{ex:filed} might be formalized with a dynamic law that has $u(married)$ as its consequence.
Law (\ref{eq:state-constraint}) is called \emph{state constraint} and says that, in any state in which $l_1, \ldots, l_n$ hold, $l_0$ also holds. 
As in $\mathcal{AL}$, state constraints allow for an elegant and concise representation of the indirect effects of actions, increasing the expressive power of the language significantly.
Law (\ref{eq:exec-condition}) is called \emph{executability condition} and intuitively says that $e$ cannot be executed if $l_1, \ldots, l_n$ hold.
A set of laws of $\mathcal{AL}_{I\!R}$ is called \emph{action description}.\footnote{Technically speaking, a set of laws of $\mathcal{AL}_{I\!R}$ should always be accompanied by a specification of a signature. For simplicity, we omit the signature whenever possible and infer it from the statements.} The semantics of $\mathcal{AL}_{I\!R}$ maps action descriptions to transition diagrams, as described next.

A set $S$ of  extended literals is \emph{closed under a state constraint} (\ref{eq:state-constraint}) if $\{ l_1,  \ldots,  l_n \} \not\subseteq S$ or $l_0 \in S$.
A set $S$ of extended literals is \emph{consistent} if, for every $f \in \mathcal{F}$, at most one of $\{ f, \neg f, u(f) \}$ is in $S$. It is \emph{complete} if at least one of $\{ f, \neg f, u(f) \}$ is in $S$.
A \emph{state} of an action description $AD$ is a complete and consistent set of extended literals closed under the state constraints of $AD$.

Given an action $a$ and a state $\sigma$, the set of \emph{(direct) effects of $a$ in $\sigma$}, denoted by $E(a,\sigma)$, is the set that contains an extended literal $\extlit$ for every dynamic law (\ref{eq:dynamic-law}) s.t. $\{ l_1, \ldots, l_n\} \subseteq \sigma$ and $e \in a$.

Consider $A=\{ A_1, A_2, \ldots, A_k \}$ where every $A_i$ is a set of extended literals. Let $B$ be a set of extended literals. We define 
%
$A \cross B=\{ A_i \cup \{b\} \,|\, A_i \in A, b \in B \}$.
For instance:
\begin{itemize}
\item
$
\{ \{ p \}, \{ q \} \} \cross \{ r, \neg r \} = \{ \{ p, r \}, \{ p, \neg r \}, \{ q, r \}, \{\ q, \neg r \} \}
$
\item
$
\{ \{ p, q \} \} \cross \{r, \neg r\} \cross \{s, \neg s \} = \{ \{ p, q, r, s\}, \{p, q, r, \neg s\}, \{p, q, \neg r, s\}, \{p, q, \neg r, \neg s \} \}
$
\end{itemize}
\begin{definition}
Let $a$ be an action and $\sigma$ be a state. The \emph{expansion of $E(a,\sigma)$} is:
\[
\mathbb{E}(a,\sigma)=\{ E(a,\sigma) \cap Lit \} \cross \{ f_1, \neg f_1, u(f_1) \} \cross \ldots \cross \{ f_k, \neg f_k, u(f_k) \}
\]
where $\{ f_1, \ldots, f_k \}$ is the set of fluents such that $u(f_i) \in E(a,\sigma)$ for every $1 \leq i \leq k$.
\end{definition}

(For an illustration of the notion of expansion, refer to Example \ref{ex:transition} below.) Given a set $S$ of extended literals and a set $Z$ of state constraints, the \emph{set, $Cn_Z(S)$, of consequences of $S$ under $Z$} is the smallest set of extended literals that contains $S$ and is closed under $Z$.
Finally, an action $a$ is \emph{executable} in a state $\sigma$ if there is no executability condition (\ref{eq:exec-condition}) such that $\{ l_1,  \ldots, l_n \} \subseteq \sigma$ and $e \in a$.

The semantics of an action description $AD$ is defined by its \emph{transition diagram} $\tau(AD)$, i.e., a directed graph $\tbeg N, E \tend$ such that:
\begin{itemize}
\item
$N$ is the collection of all states of $AD$, and
\item
$E$ is the set of all triples $\tbeg \sigma, a, \sigma' \tend$ where $\sigma$,  $\sigma'$ are states, $a$ is an action executable in $\sigma$, and $\sigma'$ satisfies the \emph{expanded successor-state equation}:
\begin{equation}\label{eq:expanded-successor-state}
\sigma'=Cn_Z(W \cup (\sigma \cap \sigma')) \mbox{ for some } W \in \mathbb{E}(a,\sigma).
\end{equation}
\end{itemize}
As before, $Z$ is the set of all state constraints of $AD$.  The argument of $Cn_Z$ in (\ref{eq:expanded-successor-state}) is the union of (i) the set of direct effects $E(e,\sigma)$ for each $e \in a$ with (ii) the set $\sigma\cap\sigma'$ of the facts ``preserved by inertia''.  The application of $Cn_Z$ adds the ``indirect effects'' to this union.
Triple $\tbeg \sigma, a,\sigma' \tend$ is called a \emph{transition} of $\tau(AD)$ and $\sigma'$ is a \emph{successor state of $\sigma$} (under $a$). A \emph{path} in a transition diagram $\mathcal{T}(AD)$ is a sequence $\pi=\tbeg \sigma_0, a_0, \sigma_1, a_1, \sigma_2, \ldots, \sigma_n \tend$ in which every triple $\tbeg \sigma_i, a_i, \sigma_{i+1} \tend$ satisfies the expanded successor state equation. We denote the initial  state of a path $\pi$  by $\pi_{\sigma_0}$.

\begin{example}\label{ex:transition}
{\em Consider an action description $\{ e_1 \causes f_1; e_1 \causes u(f_2); f_3 \lif f_1 \}$, a state $\sigma_0=\{ \neg f_1, \neg f_2,$ $\neg f_3 \}$, and action $a_0=\{ e_1 \}$. One can check that $E(a_0,\sigma_0)=\{ f_1, u(f_2) \}$. Note the lack of knowledge about $f_2$, due to the non-deterministic nature of $e_1$. Our definition of transition incorporates the idea that, in the presence of uncertainty about the effects of an action, one may sometimes  model that uncertainty explicitly by means of $u(f)$ literals, and sometimes reason by cases by ``replacing'' $u(f)$ by $f$ and $\neg f$. The set $\mathbb{E}(a_0,\sigma_0)$ captures this intuition, yielding the three possible options $\{ f_1, u(f_2) \}$, $\{ f_1, f_2 \}$, and $\{ f_1, \neg f_2 \}$. Through (\ref{eq:expanded-successor-state}), each option leads to a different successor state, $\{ f_1, u(f_2), f_3 \}$, $\{ f_1, f_2, f_3 \}$, and $\{ f_1, \neg f_2, f_3 \}$, obtained by taking into account the consequences of state constraints. Figure \ref{fig:trans-diagram} (see \ref{appendix:images}) illustrates the corresponding transitions.}
\end{example}

Intuitively, the first state from Example \ref{ex:transition} is the most ``economical,'' in that it is obtained with the least amount of assumptions, while the other two are less ``economical.'' To keep track of where reasoning by cases is applied, we introduce the following definition.
\begin{definition}
The \emph{branching-set} of a transition $\tbeg \sigma, a, \sigma' \tend$ is:
\[
\beta(\tbeg \sigma, a, \sigma' \tend)=\{ f \,|\, u(f) \in E(a,\sigma) \mbox{ and } u(f) \not\in \sigma' \}
\]
\end{definition}
For Example \ref{ex:transition}, the branching-set for the first successor state considered is $\emptyset$, while for the other two it is $\{ f_2 \}$, indicating reasoning by cases over $f_2$.

We call an action description $AD$ \emph{non-deterministic} when multiple successor states exist for a given $\sigma$ and $a$. 
%
%
Furthermore, $AD$ has \emph{emergent non-deterministic behavior} if, for some $a$ and $\sigma$, there exist multiple $\sigma'$ such that the following equation \citep{bg00} is satisfied:
\begin{equation}\label{eq:classical-successor-state}
\sigma'=Cn_Z(E(a,\sigma) \cup ( \sigma \cap \sigma')).
\end{equation}
In this paper, we focus on action descriptions without emergent non-deterministic behavior.\footnote{Action description $\{ q \lif \neg r, p;\ r \lif \neg q,  p;  a \causes p \}$ has an emergent non-deterministic behavior.
}

Next, we turn our attention to the  use of transition diagrams to reason about sequences of actions and to determine the relevance of available sources. 
\section{Reasoning about Relevance of Sources}\label{sec:similarity}
In order to enable reasoning about the relevance of sources, we begin by formalizing the notions of source and query. A source is a tuple $\tbeg \Psi, \mathcal{D},AD, I, \aleph \tend$, where $\Psi$ is a signature, $\mathcal{D}$ is a (possibly empty) subset of $\mathcal{F}$ (called the set of \emph{default fluents}), $AD$ is an action description, $I$ is a consistent set of fluent literals (intuitively capturing the available information about the initial state of the source), and $\aleph=\tbeg a_0,   a_1, \ldots, a_k \tend$ is a sequence of actions (which occur in the source).  A query $\query$ is a fluent. Intuitively, default fluents are fluents that should be assumed to be false by default at the beginning of a source.

In our approach, a \emph{qualified action sequence} is a tuple $s=\tbeg a_0/q_0,$ $a_1/q_1, \ldots,$ $a_k/q_k \tend$ where each $a_i$ is an action and each $q_i$ is a set of fluents, called \emph{qualifier}. Intuitively, a qualifier specifies to which effects of the corresponding action one should apply reasoning by cases. In reference to Example \ref{ex:transition}, the expression $e_1/\{f_2\}$ intuitively means that the reasoner should consider the transitions in which $f_2$ and $\neg f_2$ hold in the successor state, while $e_1/\{\}$ indicates that only the transition resulting in $u(f_2)$ should be considered.
The \emph{length of $s$} is $k+1$.\ The \emph{branching degree of $s$} is $\Delta(s)=|q_0| + |q_1| + \cdots + |q_k|$. Given a sequence of actions $\aleph=\tbeg a_0,   a_1, \ldots, a_k \tend$, we say that $s=\tbeg a_0/q_0, a_1/q_1, \ldots a_k/q_k \tend$ \emph{extends} $\aleph$ for every possible choice of qualifiers. $\aleph^?$ denotes the extension where all qualifiers are $\{ \}$ and $\aleph^\times$ denotes the extension where all are $\mathcal{F}$.
Let $\sigma$ be a state and $s$ be a qualified action sequence. A path $\pi=\tbeg \sigma_0,$ $a_0,$ $\sigma_1,$ $\ldots,$ $a_{k},$ $\sigma_{k+1} \tend$ is a \emph{model of $[ \sigma$, $s]$} if
(a) $\sigma_0 = \sigma$, and
(b) $\beta(\tbeg \sigma_i, a_i, \sigma_{i+1} \tend)=q_i$.
Given a set $\Sigma$ of states and a qualified action sequence $s$, a path $\pi$ is a \emph{model of $[ \Sigma$,  $s ]$} if $\pi$ is a model of $[ \sigma$, $s ]$ for some $\sigma \in \Sigma$. 
Consider an action description
$\{ a_1 \causes \neg g \lif g;$
$a_2 \causes u(f)$ $\lif \neg g \}$.
Let $\sigma$ be $\{ \neg f, g \}$. One can check that $s_1=[ \sigma, \tbeg a_1/\emptyset, a_2/\emptyset \tend ]$ has a unique model, $\tbeg \{ \neg f, g \},$ $a_1,$ $\{ \neg f, \neg g \},$ $a_2,$ $\{ u(f), \neg g \} \tend$. On the other hand, $s_{2}=[ \sigma, \tbeg a_1/\emptyset, a_2/\{f\} \tend ]$ has two models, $\tbeg \{ \neg f, g \},$ $a_1,$ $\{ \neg f, \neg g \},$ $a_2,$ $\{ f, \neg g \} \tend$ and $\tbeg \{ \neg f, g \}, a_1, \{ \neg f, \neg g \},$ $a_2, \{ \neg f, \neg g \} \tend$. Hence, $\Delta(s_1)=0$ and $\Delta(s_2)=|\emptyset|+|\{f\}|=1$.

Now we turn our attention to incomplete knowledge about the initial state. In our approach, the \emph{default assumption} is to consider the default fluents false and to assume that  $u(f)$ holds for every non-default fluent $f$. However, as highlighted by Section \ref{sec:analysis}, commonsense sometimes leads one to explore cases beyond those of the default assumption -- either considering that a default fluent might be true, or reasoning by cases over the truth of the other fluents. 

This intuition is captured by the notion of \emph{forcing of a fluent}. A fluent whose truth value is currently unknown is \emph{forced} when a reasoner wants to consider for it cases other than those from the default assumption. More precisely, let $I$ be a consistent set of extended literals and $f$ be a fluent that should be forced. The \emph{forcing of $f$ in $I$}, written $I[f]$, is defined as follows:
\[
I[f]=\left\{
\begin{array}{rl}
\{ I \cup \{ f \} \} & \mbox{if $f \in \mathcal{D}$ and $\{ \neg f,  u(f) \} \cap I = \emptyset$} \\
\{\ I \cup \{ f \},\ I \cup \{ \neg f \} \ \} & \mbox{if $f \not\in \mathcal{D}$ and $\{ f, \neg f,  u(f) \} \cap I = \emptyset$}\\
\{I\} &\ \mbox{otherwise}
\end{array}
\right.
\]
For an example of the notion of forcing, consider the following.
\begin{example}\label{ex:date-forcing}
{\em
Consider $\mathcal{S}_1$ from Example \ref{ex:date}, ``John went on his first date with Mary.'' The source is encoded by tuple $\tbeg \Psi, \mathcal{D}, AD, I, \aleph \tend$ where $\Psi=\tbeg \mathcal{F}, \mathcal{D}, \mathcal{E} \tend$ and:
\begin{itemize}
\item
$\mathcal{F}$ consists of fluents: $m$ -- John is married\footnote{In practical cases, one will want to introduce variables to increase generality, e.g. $m(X)$ for $X$ is married.}; $ab$ -- John is an exception w.r.t. going on  first dates when  married.
\item
$\mathcal{D}=\{ab\}$, i.e. by default, John is not an exception.
\item
$\mathcal{E}$ consists of action $d$, i.e. going on a first date.
\item
Action description $AD$ consists of law $\{ \mbox{impossible } d \mbox{ if } m, \neg ab \}$, intuitively stating that a married person does not \emph{normally} go on first dates.\footnote{Note the use of default fluent $ab$ to formalize the fact that action $d$  is \emph{normally} impossible if one is married.}
\item
The knowledge about the initial state of the source is captured by $I=\emptyset$.
\item
The sequence of actions is $\aleph=\tbeg d \tend$, i.e. John went on a first date.
\end{itemize}
Finally, query $\query$ is given by fluent $m$.  Because the story does not say whether $m$ holds in the initial state, the forcing of $m$ in $I$ allows one to consider both cases explicitly. One can check that $I[m]$ is $\{ \{ m \}, \{ \neg m \} \}$.
}
\end{example}
It is important to note that fluents that already occur in $I$ are not affected by the forcing. The notion of forcing is extended to sets of fluents in a natural way. The \emph{forcing of $\{ f_1,$$\ldots,$$f_m \}$ in $I$} is defined recursively as follows:
\[
I[\{f_1,\ldots,f_m\}]=\left\{
\begin{array}{rl}
I[f_1] & \mbox{if $m=1$} \\
\{ I'[f_m] \,\,|\,\, I' \in I[\{ f_1,  \ldots, f_{m-1} \}] \} & \mbox{if $m>1$}
\end{array}
\right.
\]
%
In the case of Example \ref{ex:date-forcing}, one can check that $I[m,ab]$ is $\{ \{ m, ab \}, \{ \neg m, ab \} \}$, intuitively meaning that both cases are considered for $m$ and that default fluent $ab$ is hypothesized to be true.

Once the fluents that deserve special treatment have been handled, the default assumption is applied to all remaining fluents whose truth value is still unknown. This process is captured by the notion of completion, defined next.
\begin{definition}\label{def:completion-aux}
Let $I$ be a consistent set of fluent literals and  $Z$ be the set of state constraints of an action description $AD$. The \emph{completion of $I$}, denoted by $\gamma(I)$, is the consistent set of extended literals obtained as follows:
\begin{enumerate}[leftmargin=.31in]
\item Let $I'$ be obtained by expanding $I$ with a fluent literal $\neg f$ for every default fluent $f$ such that $f \not\in I$.
\item
If $Cn_Z(I')$ is consistent, then $\gamma(I)$ is the union of $Cn_Z(I')$ and a literal $u(f)$ for every $f$ that does not occur in $Cn_Z(I')$. Otherwise, $\gamma(I)$ does not exist.
\end{enumerate}
\end{definition}
For an example of a case in which $\gamma(I)$ does not exist, consider $I=\{p, q\}$, $\mathcal{D}=\emptyset$ and $AD=\{ \neg q \mbox{ if } p \}$. Given that there are no default fluents, $I'=I$. The application of $Cn_Z$ to $I'$ results in $\{ p, q, \neg q \}$, which is inconsistent. Hence, the completion of $I$ does not exist.

When $\gamma(I)$ exists, it is not difficult to check that the following holds:
\begin{proposition}\label{prop-1}
For any consistent set of fluent literals $I$, $\gamma(I)$ is complete,  consistent and includes $I$. 
\end{proposition}
%
We can now combine forcing of a set of fluents $F$ and completion of its outcomes as follows:
\begin{definition}\label{def:completion}
Let $F$ be a set of fluents. The \emph{completion of $I$ w.r.t. $F$} is the set:
\[
\gamma(I,F)=\{ \gamma(I') \,|\, I' \in I[F] \mbox{ and } \gamma(I') \mbox{ exists} \}.
\]
The \emph{degree of $\gamma(I,F)$}, denoted by $\Delta(\gamma(I,F))$, is $|F|$.
\end{definition}
The following example illustrates this concept.
\begin{example}\label{ex:completion}
{\em
Continuing Example \ref{ex:date-forcing}, let us find $\gamma(I,F)$ for $F=\{m\}$. Intuitively, this means that we would like to consider explicitly the possible options for the truth value of $m$, while applying the default assumption to all other fluents.

According to Definition \ref{def:completion}, first we need to find the forcing of $F$ in $I$. By definition of forcing, $I[F]=I[m]$. In Example \ref{ex:date-forcing}, we found $I[m]$ to be $I[m]=\{ \{ m \}, \{ \neg m \} \}$.  Hence, $I[F]=\{ \{ m \}, \{ \neg m \} \}$. 

Next, we apply the default assumption to every $I' \in I[F]$ by finding $\gamma(I')$. From Definition \ref{def:completion-aux} it follows that $\gamma(\{m\})=\{m,\neg ab\}$ and $\gamma(\{\neg m\})=\{\neg m,\neg ab\}$. Hence, $\gamma(I,F)= \{ \{ m, \neg ab\}, \{ \neg m,  \neg ab \} \}$. Intuitively, this corresponds to a situation in which a reasoner considering possible initial states for the scenario makes the default assumption about default fluent $ab$, but decides to reason by cases about $m$.
}
\end{example}

As demonstrated by Example \ref{ex:date}, there are cases in which the truth of certain fluents in the initial state can be inferred indirectly from the source. By building on the previous definitions, we can now make this idea precise in the following. 
\begin{definition}\label{def:expansion}
Given a consistent set $I$ of fluent literals and a sequence of actions $\aleph$, the \emph{conservative expansion of $I$ under a $\aleph$} is:
\[
\ifShortVersion
\begin{small}
\fi
\varepsilon(I,\aleph)=\bigcap_{I' \in I[\mathcal{F} \setminus \mathcal{D}]} \{ I' \,|\, [ \gamma(I'), \aleph^\times ] \mbox{ has a model} \}
\ifShortVersion
\end{small}
\fi
\]
\end{definition}
The intuition behind this definition is that the reasoner expands $I$ by considering all possible cases for the non-default fluents from $I$. For each resulting expanded set $I'$,  the reasoner checks if there exists a completion $\gamma(I')$ compatible with the actions in $\aleph$. The conservative expansion of $I$ under $\aleph$ consists of the extended literals shared by all expanded sets $I'$ satisfying this test. To ensure that all possible contingencies are considered, the definition applies reasoning by cases to the effects of the actions in $\aleph$ -- hence the use of $\aleph^\times$. 

The above definition yields a number of important properties.
\begin{proposition}\label{prop-2}
\begin{enumerate}[leftmargin=.31in]
\item
If $\gamma(I')$ does not exist for any element of $I[\mathcal{F}\setminus \mathcal{D}]$, then $\varepsilon(I,\aleph)$ does not exist. 
\item
If $\varepsilon(I,\aleph)$ exists, then $I \subseteq \varepsilon(I,\aleph)$.
\item
When it exists, $\varepsilon(I,\aleph)$ is consistent but not necessarily complete.
\end{enumerate}
\end{proposition}
Note that, if $\varepsilon(I,\aleph)$ does not exist, this intuitively indicates that there is some fundamental inconsistency in the story and the source should be considered irrelevant to any query.  As we will see later, this is handled by assigning a semantic score of $\infty$ to the source.
\begin{example}
{\em
Let us calculate $\varepsilon(I,\aleph)$ for our running example. Recall that $\aleph=\tbeg d \tend$. The first step consists in checking for models of $[ \gamma(I'), \aleph^\times ]$ where $I' \in I[\mathcal{F} \setminus \mathcal{D}]$. From Example \ref{ex:completion}, we know that $I[\mathcal{F} \setminus \mathcal{D}]=\{ \{ m \}, \{ \neg m \} \}$ and that the completions of each component of the set are, respectively, $\{m,\neg ab\}$ and $\{\neg m,\neg ab\}$. We can now check for models. Clearly, $[ \{ m, \neg ab\}, \tbeg d \tend ]$ has no model, because $d$ is not executable in $\{ m, \neg ab \}$. On the other hand, $[ \{ \neg m, \neg ab\}, \tbeg d \tend ]$ has a model. The second step consists in calculating the intersection of all $I'$ that satisfy the requirements. In this example, $\varepsilon(I,\aleph)$ is the intersection of the only set $\{ \neg m \}$, resulting in $\varepsilon(I,\aleph)=\{ \neg m\}$. That is, we have inferred that John is not married in the initial state, which is aligned with the conclusion reached in Example \ref{ex:date}.
}
\end{example}
We are now ready to introduce the notion of entailment and to use it to determine whether there is a match between $\query$ and $\mathcal{S}$. A path $\pi=\tbeg \sigma_0, a_0, \sigma_1,  \ldots, a_{k-1}, \sigma_{k} \tend$ \emph{entails} a fluent literal $l$ (written $\pi \models l$) if $l \in \sigma_k$. Given a fluent $f$, we say that \emph{$\pi$ entails $\pm f$} (written $\pi \models \pm f$) if $\pi \models f$ or $\pi \models \neg f$. The main notion of this section is given next.
\begin{definition}\label{def:match}
Given a source $\mathcal{S}=\tbeg \Psi, AD, I, \aleph \tend$ and a query $\query$, we say that \emph{$\mathcal{S}$ is a match for $\query$} if there exist a set $F$ of fluents from $\Psi$ and a qualified action sequence $s$ extending $\aleph$ such that:
\begin{enumerate}[label=\textbf{c\arabic*},leftmargin=.5in]
\item\label{crit:1}
$\pi \entails \pm \query$ for some model $\pi$ of $[ \gamma(\varepsilon(I,\aleph),F), s ]$, and
\item\label{crit:2}
for every model $\pi'$ of $[ \gamma(\pi_{\sigma_0} \setminus \varepsilon(I,\aleph),\emptyset)$, $\tbeg \  \tend ]$ , one of the following holds:
\begin{enumerate}[leftmargin=.31in]
\item\label{crit:2a}
$\pi' \not\entails \pm \query$, or
\item
$\pi' \entails \neg \query$ and $\pi \entails \query$, or 
\item
$\pi' \entails \query$ and $\pi \entails \neg \query$.
\end{enumerate}
\end{enumerate}
\end{definition}
Condition (\ref{crit:1}) checks whether the $\mathcal{S}$ is relevant to $\query$, that is, if it has any bearing on the truth value of $\query$. To do so, the reasoner is allowed to reason by cases about an arbitrary set of non-default fluents and to assume that some default fluents behave exceptionally. This choice of fluents is embodied by set $F$ and by its role in (\ref{crit:1}). The reasoner is also allowed to reason by cases about the effects of an arbitrary set of actions from $\aleph$, as outlined by the freedom in selecting its extension $s$. Note that the key criterion that a path needs to satisfy in (\ref{crit:1}) is the entailment of $\pm\query$. 

Condition (\ref{crit:2}) ensures that the assumptions made by (\ref{crit:1}) leading to the selection of path $\pi$,  are not directly and solely responsible for the entailment of $\pm\query$. (Refer to Example \ref{ex:S1-ending} for an illustration of the application and interplay of (\ref{crit:1}) and (\ref{crit:2}).) To achieve this, given path $\pi$, condition (\ref{crit:2}) identifies the fluents of $\pi_{\sigma_0}$ that are in $I$ and those whose truth value was inferred according to Definition \ref{def:expansion}. The default assumption is applied to all of those fluents. Next, the truth value of $\query$ is checked in the state obtained in this way. If the truth value of $\query$ in this state coincides with the truth value of $\query$ in $\pi_{\sigma_0}$, this indicates that the entailment of $\pm\query$ in (\ref{crit:1}) was due solely to the assumptions made and was independent from the initial state information and actions from $\mathcal{S}$. Therefore, the current model $\pi$ should be discarded, as insufficient evidence that $\mathcal{S}$ is a match for $\query$. $\mathcal{S}$ is considered to be a match for $\query$ only when a path satisfying (\ref{crit:1}) is found and when is shown that the information from $\mathcal{S}$ is critical in entailing $\pm\query$ in $\pi_{\sigma_0}$.

In Definition \ref{def:match}, $F$ and $s$ can be viewed as an indication of the ``strength'' of the match. If a match can be found for $F=\emptyset$ and $s=\aleph^?$, it means that all that is needed to determine that $\mathcal{S}$ is a match for $\query$ is the information from $\mathcal{S}$. This makes for a ``strong match''. Instead, if a match is found only for other values of $F$ and $s$, it means that the match depends on additional assumptions, such as assuming that a default fluent is unexpectedly true or that a non-default fluent has a specific truth value. This makes for a ``weaker'' match, since such assumptions may or may not be true in reality, and will have to be checked by a reader with human-level intelligence, as discussed in Section \ref{sec:intro}. The notion of semantic score, defined next, makes this idea precise.
\begin{definition}\label{def:score}
Given a source $\mathcal{S}$ and a query $\query$, the \emph{semantic score of $\mathcal{S}$} (w.r.t. $\query$) is the smallest value of $\Delta(\gamma(\varepsilon(I,\aleph),F))+\Delta(s)$ for all possible choices of $F$ and $s$ satisfying conditions (\ref{crit:1}) and (\ref{crit:2}) from Definition \ref{def:match}. If $\mathcal{S}$ is not a match for $\query$, its semantic score is $\infty$.
\end{definition}
Note that a semantic score of $\infty$ indicates that $\mathcal{S}$ is irrelevant to the query. Definitions \ref{def:match} and \ref{def:score} provide a complete definition of the reasoning task of Action-Centered \ac{IR}. Given a query and a set of sources, the sources relevant to $\query$ can be identified by means of Definition \ref{def:match} and then ranked by relevance according to the semantic score given by Definition \ref{def:score}.
We illustrate the process by means of the following examples.

\begin{example}\label{ex:S1-ending}
{\em
Continuing our running example, let us apply Definition \ref{def:match} to check if $\mathcal{S}_1$ is a match for $\query$. Let us first look for $F$ and $s$ satisfying (\ref{crit:1}). We begin with $F=\emptyset$, $s=\tbeg d \tend^?$. We need to find a model $\pi$ of $[ \gamma(\varepsilon(I,\aleph),F), s ]$ such that $\pi \entails \pm \query$. Using the results from the previous examples, one can check that $\gamma(\varepsilon(I,\aleph),F)=$ $\gamma(\{\neg m\},\emptyset)=$ $\{ \{ \neg m,  \neg ab \} \}$ and that $[\{  \{ \neg m,  \neg ab \}\}, \tbeg d \tend^? ]$ has a unique model $\pi=\tbeg \{ \neg m, \neg ab \},$ $ d, \{ \neg m,  \neg ab\} \tend$. Thus, the model entails $\pm \query$, which means that condition (\ref{crit:1}) is satisfied.

Next, we check condition (\ref{crit:2}).  Clearly, $\gamma(\pi_{\sigma_0} \setminus \varepsilon(I,\aleph),\emptyset)=\{ \{ u(m), \neg ab \} \}$ and $[ \{ \{ u(m),$ $\neg ab \} \},$ $ \tbeg \tend ]$ has a unique model $\tbeg \{ u(m), \neg ab \} \tend$. The model does not entail $\pm \query$, and thus conditions (\ref{crit:2a}) and (\ref{crit:2}) are satisfied. Intuitively, this means that any assumptions made to satisfy (\ref{crit:1}) \emph{are not directly and solely responsible} for the ability of $\pi$ to entail $\pm\query$ in (\ref{crit:1}). Hence, it is acceptable to conclude that $\mathcal{S}$ matches $\query$. Additionally, according to Definition \ref{def:score}, the semantic score of $\mathcal{S}_1$ is $\Delta(\gamma(\varepsilon(I,\aleph),F))+\Delta(s)=|F|+\Delta(s)=|\emptyset|+\Delta(\tbeg d \tend^?)=0$.
}
\end{example}
\begin{example}
{\em
As an additional example, consider $\mathcal{S}_2$, ``John read a book,'' from Example \ref{ex:date}. As above, $\query=m$, $\mathcal{F}=\{m,ab\}$, $\mathcal{D}=\{ ab\}$, $I=\emptyset$. $\mathcal{E}$ is expanded by an additional action $r$, representing reading a book. The sequence of actions is captured by $\aleph=\tbeg r \tend$. $AD$ is as before.\footnote{For simplicity, we formalize $r$ as a no-op action.} 

Let us begin by finding the conservative expansion $\varepsilon(I,\aleph)$ through Definition \ref{def:expansion}. Similarly to the running example, $I[\mathcal{F} \setminus \mathcal{D}]$ is $\{ \{ m \}, \{ \neg m \} \}$, and its elements yield completions $\{ m, \neg ab\}$ and $\{ \neg m,  \neg ab \}$ respectively. Differently from Example \ref{ex:S1-ending}, both $[ \{\{ m, \neg ab\}\}, \tbeg\ r  \tend^\times ]$ and $[ \{\{ \neg m, \neg ab\}\},$ $ \tbeg\ r \tend^\times ]$ have models. Hence, $\varepsilon(I,\aleph)=\{ m \} \cap \{ \neg m \}=\emptyset$. That is, the story does not allow one to infer the truth value of any additional fluent. 

Next, we apply Definition \ref{def:match}. We begin by considering the models of $[ \gamma(\varepsilon(I,\aleph),F),s ]$ for $F=\emptyset$, $s=\tbeg r \tend^?$. One can check that $\gamma(\varepsilon(I,\aleph),F)=$ $\gamma(\emptyset,\emptyset) =$ $\{ \{ u(m),  \neg ab \} \}$. $[ \{\{ u(m),  \neg ab \}\}, \tbeg r \tend^? ]$ has a unique model $\pi=\tbeg \{ u(m), \neg ab \},$ $r,$ $\{ u(m),  \neg ab\} \tend$. Thus, $\pi \not\models \pm \query$. 

Another possible choice for $F$ and $s$ is $F=\emptyset$, $s=\tbeg r \tend^\times$. One can check that $[ \gamma(\varepsilon(I,\aleph),F), s ]$ has two models -- for instance, $\tbeg \{ u(m), \neg ab \}, r, \{ u(m),  \neg ab\} \tend$ -- but neither entails $\pm \query$. 

A more interesting choice is $F=\{ m \}$, $s=\tbeg r \tend^?$. Clearly, $\gamma(\varepsilon(I,\aleph),F)=\gamma(\emptyset,\{m\})$, from which it follows that $[ \gamma(\varepsilon(I,\aleph),F),s ]$ has two models: $\pi_{1}=\tbeg \{ m, \neg ab \}, r, \{ m,  \neg ab\} \tend$ and $\pi_2 = \tbeg \{ \neg m, \neg ab \}, r,$\ $ \{ \neg m,  \neg ab\} \tend$, with $\pi_1 \models \query$ and $\pi_2 \models \neg \query$ respectively. Next, we need to check condition (\ref{crit:2}) for each. For the former, $\gamma((\pi_1)_{\sigma_0} \setminus \emptyset, \emptyset)=\{ \{ m, \neg ab \} \}$, and $[ \{ \{m, \neg ab\}\ \}, \tbeg \tend ]$ has a unique model $\tbeg \{ m, \neg ab \} \tend$, which entails $\query$. Since $\query$ is also entailed by $\pi_1$, (\ref{crit:2}) is not satisfied. For  $\pi_{2}$, we obtain a unique model $\tbeg \{\{\ \neg m, \neg ab \}\} \tend$, which entails $\neg \query$, thus failing to satisfy (\ref{crit:2}) as  well. Therefore, none of these choices for $F$ and $s$ yields a match. 
%
Similar conclusions can be drawn for the other choices for $F$ and $s$. Hence, $\mathcal{S}_2$ does not match $\query$. By Definition \ref{def:score}, $\mathcal{S}_2$ has a semantic score of $\infty$.
}
\end{example}
The other examples are solved similarly. We provide highlights of their solutions.

\noindent
\uline{Example \ref{ex:date-mormon}.} In this example, people from countries that allow plural marriage are exceptions to the custom  about first dates,  and thus $I=\{ ab \}$, $\aleph=\tbeg d \tend$, and $I[\mathcal{F} \setminus \mathcal{D}]=\{ \{ m,  ab \}, \{ \neg m,  ab \} \}$. Differently from  the previous case, both sets of $I[\mathcal{F} \setminus \mathcal{D}]$ yield a model, since $ab$ makes the executability condition inapplicable. Hence, $\varepsilon(I,\aleph)=\{ ab \}$. Selecting $F=\emptyset$, $s=\tbeg d \tend^?$ yields a unique model $\tbeg \{ u(m),ab \},  d ,$ $\{ u(m),ab\} \tend \not\models \pm  \query$. Selecting $F=\{m\}$, $s=\tbeg d \tend^?$ yields two models entailing $\query$ and $\neg \query$ respectively, but the same conclusions are entailed by $[ \gamma(\pi_{\sigma_0} \setminus \varepsilon(I,\aleph),\emptyset), \tbeg \tend ]$, thus failing to satisfy condition (\ref{crit:2}). Similar reasoning applies to the other cases. Because no $F$ and $s$ satisfying Definition \ref{def:match} exist, the semantic score of $\mathcal{S}$ is $\infty$, indicating that it is irrelevant to $\query$. Note the key role played by condition (\ref{crit:2}) in this example: without it, the source would have been deemed relevant to the query.

\noindent
\uline{Example \ref{ex:filed}.} In this example, $AD$ is expanded with $w \causes m;$ $fd \causes u(m)$ and relevant executability conditions. The signature is extended accordingly. Also, $I=\emptyset$ and $\aleph=\tbeg d, w,  fd  \tend$. 

Similarly to Example \ref{ex:date}, one can check that   $\varepsilon(I,\aleph)=\{ \neg m\}$. The model obtained from $F=\emptyset$ and $s=\aleph^?$ does not entail $\pm \query$. On the other hand, the choice of $F=\emptyset$ and $s=\tbeg d/\emptyset, w/\emptyset, fd/\{m\} \tend$ yields two models, entailing $\query$ and $\neg \query$ respectively, depending on whether $m$ is true or false after $fd$. This time, condition (\ref{crit:2}) is satisfied, since, in both cases, $\gamma(\pi_{\sigma_0} \setminus \varepsilon(I,\aleph),\emptyset)=\{ \{ u(m), \neg ab \} \}$ and $[ \{ \{u(m), \neg ab\}\ \}, \tbeg \tend]$ does not entail $\pm \query$. In conclusion, $\mathcal{S}$ indeed matches $\query$. In this case, the source has semantic score of $1$, which is, as one might expect, worse than that of $\mathcal{S}_1$ from Example \ref{ex:date}, while better than that of $\mathcal{S}_2$.

\ifLongVersion
\textcolor{red}{PARTS THAT MUST BE EXTENDED ARE BELOW.}

\textbf{Handling probabilistic assumptions.} Let $\lambda(l)$ be the probability of $l$ being true, where $l$ true is the assumption we made. Such a probability could be specified as part of the knowledge base/action description. Then, the probability of a path $\pi$ is $\lambda(\pi)=\prod_{l \in assumptions(\pi)} \lambda(l)$.

\textbf{A better estimate of relevance.} We should weigh the current value of the rank of $\mathcal{S}$ by the fraction of paths corresponding to $F$ and $s$ that yield $\pm \query$.

\textbf{An estimate of the accuracy of the match.} I must find my diagram with $\alpha$ in it!!!

\fi

\section{Automating the Reasoning Task}\label{sec:implementation}
In this section, we propose an approach for automating Action-Centered \ac{IR}. Our approach leverages a translation of $\mathcal{AL}_{I\!R}$ to ASP. 
The choice of ASP is motivated by the availability of well-understood mappings from action language $\mathcal{AL}$ and its semantics to ASP, as well as of ASP-based implementations of the other modeling and reasoning techniques, discussed earlier, which we build upon. Additionally, ASP's non-monotonic nature allows one to model, in a natural and declarative way, crucial elements such as the effects of non-deterministic actions and the different ways of formalizing uncertainty (by cases vs. explicit lack of knowledge). A brief introduction on ASP can be found in \ref{sec:ASP}.

\subsection{ASP Implementation of the Reasoning Task}\label{sec:algorithm}
Given  a consistent set $I$ of fluent literals, a set $F$ of fluents, a qualified action sequence $s$, and an action description $AD$, the encoding of $\mathcal{AL}_{I\!R}$ is program $\Pi_{AD}(I,F,s)$, described next.

In the following, $\mathtt{I}$ ranges over steps in the evolution of the domain\footnote{We assume that the range of $\mathtt{I}$ is provided by the process of translating the passage to a logical representation.}; given fluent literal $l$, $\chi(l,\mathtt{I})$ stands for $holds(f,\mathtt{I})$ if $l=f$ and $\neg holds(f,\mathtt{I})$ if $l=\neg f$. The translation of a dynamic law of the form (\ref{eq:dynamic-law}) depends on the form of $\extlit$. If $\extlit$ is a fluent literal, the translation is:
\[
\chi(\extlit,\mathtt{I}+1) \hif occurs(e,\mathtt{I}), \chi(l_1,\mathtt{I}), \ldots, \chi(l_n,\mathtt{I}).
\]
If $\extlit$ is of the form $u(f)$, the translation of the law is:
\[
\ifShortVersion
\fi
\begin{array}{l}
u(f,\mathtt{I}+1) \hif occurs(e,\mathtt{I}), \chi(l_1,\mathtt{I}), \ldots, \chi(l_n,\mathtt{I}), \lpnot split(f,\mathtt{I}).\\
\chi(f,\mathtt{I}+1) \lor \chi(\neg f,\mathtt{I}+1) \hif occurs(e,\mathtt{I}), \chi(l_1,\mathtt{I}), \ldots, \chi(l_n,\mathtt{I}), split(f,\mathtt{I}).
\\
\end{array}
\ifShortVersion
\fi
\]
Expression $occurs(e,\mathtt{I})$ states that elementary action $e$ occurs at step $\mathtt{I}$ in the story; $split(f,\mathtt{I})$ indicates that reasoning by cases should be applied to fluent $f$. 

A state constraint of the form (\ref{eq:state-constraint}) is translated as an ASP\ rule of the form
\[
\chi(l_0,\mathtt{I}) \hif \chi(l_1, \mathtt{I}), \ldots, \chi(l_n,\mathtt{I}).
\]
Finally, an executability condition of the form (\ref{eq:exec-condition}) is translated as a rule
\[
\hif occurs(e,\mathtt{I}), \chi(l_1,\mathtt{I}), \ldots, \chi(l_n,\mathtt{I}).
\]
The translation is completed by a set of general-purpose axioms that formalize the semantics of $\mathcal{AL}_{I\!R}$. The following rules capture the notion of consistency of sets of fluents ($\mathtt{F}$ is a variable ranging over all fluents):
\[
\hif \chi(\mathtt{F},\mathtt{I}),u(\mathtt{F},\mathtt{I}).\ \ \ \ \ \ \ \ \ \ \ \ \ \ \ \ 
\hif \chi(\neg \mathtt{F},\mathtt{I}),u(\mathtt{F},\mathtt{I}).
\]

Next are the inertia axioms, which are expanded in $\mathcal{AL}_{I\!R}$ to accommodate extended literals:
\[
\ifShortVersion
\fi
\begin{array}{l}
\chi(\mathtt{F},\mathtt{I}+1) \hif \chi(\mathtt{F},\mathtt{I}), \lpnot \chi(\neg \mathtt{F},\mathtt{I}+1), \lpnot u(\mathtt{F},\mathtt{I}+1). \\
\chi(\neg \mathtt{F},\mathtt{I}+1) \hif \chi(\neg \mathtt{F},\mathtt{I}), \lpnot \chi(\mathtt{F},\mathtt{I}+1), \lpnot u(\mathtt{F},\mathtt{I}+1).\\
u(\mathtt{F},\mathtt{I}+1) \hif u(\mathtt{F},\mathtt{I}), \lpnot \chi(\mathtt{F},\mathtt{I}+1), \lpnot \chi(\neg \mathtt{F},\mathtt{I}+1). \\
\end{array}
\ifShortVersion
\fi
\]
The final set of axioms captures the definition of completion: 
\[
\ifShortVersion
\fi
\begin{array}{l}
\ifLongVersion
\chi(\mathtt{F},0) \hif init(\mathtt{F}).\\
\chi(\neg \mathtt{F},0) \hif \neg init(\mathtt{F}).\\
\else
[\mathbf{g_1}]\ \ \chi(\mathtt{F},0) \hif init(\mathtt{F}).\ \ \ \ \ \ \ \ \ \ 
\chi(\neg \mathtt{F},0) \hif \neg init(\mathtt{F}).\\
\fi
[\mathbf{g_2}]\ \ \chi(\mathtt{F},0) \hif forced(\mathtt{F}), default(\mathtt{F}), \lpnot \neg init(\mathtt{F}).\\

\ \ \ \ \ \ \ \ \ \chi(\mathtt{F},0) \lor \chi(\neg \mathtt{F},0) \hif forced(\mathtt{F}), \lpnot default(\mathtt{F}), \\
\hspace*{1.46in}\lpnot init(\mathtt{F}), \lpnot \neg init(\mathtt{F}). \\

[\mathbf{g_3}]\ \ \chi(\neg \mathtt{F},0) \hif default(\mathtt{F}), \lpnot \chi(\mathtt{F},0).\\
\ \ \ \ \ \ \ \ \ u(\mathtt{F},0) \hif \lpnot default(\mathtt{F}), \lpnot \chi(\mathtt{F},0), \lpnot \chi(\neg \mathtt{F},0).\\
\end{array}
\ifShortVersion
\fi
\]
Above, atom $default(f)$ states that $f$ is a default fluent, and is included in the form of a fact, as a part of the translation, for every $f\in\mathcal{D}$. 
Atom
$init(f)$ (resp., $\neg init(f)$)
says that 
$f$ is initially true (resp.,  false), i.e. is part of set $I$.
Atom $forced(f)$ states that $f$ is forced.

Set $[\mathbf{g_1}]$ of rules maps the knowledge about the initial state to atoms of the form $holds(\cdot,\cdot)$.
Set $[\mathbf{g_2}]$ formalizes to the definition of forcing: the first rule ensures that a forced default fluent is set to true, and the second rule states that, when a non-default fluent is forced, both possible truth values should be considered for it. Set $[\mathbf{g_3}]$ applies the default assumption and follows closely Definition \ref{def:completion}: default fluents default to false, and non-default fluents default to unknown. 

The next step of the definition of $\Pi_{AD}(I,F,s)$ is the encoding of its arguments. For every $f \in I$ (resp., $\neg f \in I$), $\Pi_{AD}(I,F,s)$ includes a fact $init(f)$ (resp., $\neg init(f)$). For every $f \in F$, $\Pi_{AD}(I,F,s)$ includes a fact $forced(f)$. 
Qualified action sequence $s$ is encoded by a set of facts of the form $occurs(e,i)$ and $split(f,i)$, where every $e$ and $f$ is from $s$, and $i$ is the corresponding index in the sequence of elements from $s$. 

One can check that an expression of the form $\{e_1, \ldots, e_m \}^?$ at position $i$ of $s$ (where  each $e_i$ is an elementary action) is translated into a collection of statements $occurs(e_1,i), \ldots, occurs(e_m,i)$. An expression of the form $\{e_1, \ldots, e_m \}^\times$ at position $i$ of $s$ is translated into a collection of statements $occurs(e_1,i), \ldots, occurs(e_m,i)$ together with a statement $split(f,i)$ for every $f \in \mathcal{F}$.

This completes the definition of $\Pi_{AD}(I,F,s)$. Next, we link its answer sets to the models of $[ \gamma(I,F),s ]$.
We say that an answer set $A$ of a program \emph{encodes a path $\pi$} if:
\begin{enumerate}[leftmargin=.5in]
\item[(a)]
for every fluent literal $l$, $l \in \sigma_i$ iff $\chi(l,i) \in A$;
\item[(b)]
for every fluent $f$, $u(f) \in \sigma_i$ iff $u(f,i) \in A$;
\item[(c)]
for every elementary action $e$, $e \in a_i$ iff $occurs(e,i) \in A$;
\end{enumerate}
The link between answer sets and models is established by the following.
\begin{theorem}\label{theorem:asp-and-models}
Let $I$ be a consistent set of fluent literals, $F$ a set of fluents, and $s$ a qualified action sequence. A path $\pi$ is a model of $[ \gamma(I,F)$, $s ]$ iff there exists an answer set of $\Pi_{AD}(I,F,s)$ that encodes $\pi$.
\end{theorem}

\begin{corollary}\label{corollary:asp-and-models}
\begin{itemize}[leftmargin=.25in]
\item
A model $\pi$ of $[ \gamma(I,F)$, $s ]$ that entails a fluent literal $l$ exists iff there exists an answer set $A$ of $\Pi_{AD}(I,F,s)$ such that $\chi(l,k) \in A$, where $k$ is the length of $s$. 
\item
For every fluent $f$, $\pi \models \pm f$ iff $\{ \chi(f,k), \chi(\neg f,k) \} \cap A \not= \emptyset$.
\end{itemize}
\end{corollary}
These results motivate the $\texttt{FindMatch}$ algorithm, shown in Figure \ref{alg:FindMatch}.
Let $||A||$ be the number of atoms of $A$ formed by relations $forced$ and $split$ (with $||A||=\infty$ if $A=\bot$).
\begin{figure}[htbp]
\fbox{\parbox[c]{1.0\textwidth}{
\begin{small}
\uline{\textbf{Algorithm:} \texttt{FindMatch}($I$,$\aleph$,$\query$)}\\
\textbf{Input:} \\
\hspace*{.2in}$I$ -- set of fluent literals explicitly stated to hold in the initial state by $\mathcal{S}$;\\
\hspace*{.2in}$\aleph=\tbeg a_0, a_1, \ldots, a_k \tend$ -- sequence of actions from $\mathcal{S}$; \\
\hspace*{.2in}$\query$ -- fluent.\\
\textbf{Output:} an answer set encoding a path if a match exists; $\bot$ otherwise.
\ifShortVersion
\vspace*{-.1in}
\fi
\begin{enumerate}
\item\label{alg:match:step-find_ix}
Let $R$ be the intersection of all answer sets of $\Pi_{AD}(I,\mathcal{F} \setminus \mathcal{D},\aleph^\times)$ and $I'$ be $I \cup
\{ l \,|\, \{ \chi(l,0), forced(f) \} \subseteq R \land ( l=f \lor l=\neg f ) \}$.
\item\label{alg:match:step-existence}
If $\Pi_{AD}(I,\mathcal{F} \setminus \mathcal{D},\aleph^\times)$ has no answer set, return $\bot$ and terminate.
\item\label{alg:match:step-init} 
Initialize $F:=\emptyset$ and $s:=\aleph^?$.
\item\label{alg:match:step-find_as} 
For every answer set $A$ of $\Pi_{AD}(I',F,s)$ such that $\{ \chi(\query,k+1),$ $\chi(\neg \query,k+1) \} \cap A \not= \emptyset$:
\begin{enumerate}
\item
Let $X=\{ f \,|\,  \chi(f,0) \in A \land f \not\in I' \}
   \cup
   \{ \neg f \,|\,  \chi(\neg f,0) \in A$ $\land$ $\neg f \not\in I' \}.$
\item\label{alg:match:step-verify}
For every answer set $B$ of $\Pi_{AD}(X,\emptyset,\tbeg\ \tend)$, check that one of the following holds:
\begin{itemize}
\item
$\{ \chi(\query,0), \chi(\neg \query,0) \} \cap B = \emptyset$, or
\item
$\chi(\query,0) \in B \land \chi(\neg \query,k+1) \in A$, or
\item
$\chi(\neg \query,0) \in B \land \chi(\query,k+1) \in A$.
\end{itemize}
\item\label{alg:match:step-return}
If the test at step (\ref{alg:match:step-verify}) succeeds, then return $A$ and terminate.
\end{enumerate}
\item\label{alg:match:select}
Select a set $F'$ of fluents and an extension $s'$ of $\aleph$ such that:
\begin{enumerate}
\item
the pair $F'$, $s'$ has not yet been considered by the algorithm, and 
\item\label{alg:match:minimality}
$|F'|+\Delta(s')$ is minimal among such pairs.
\end{enumerate}
\item\label{alg:match:done}
If no such pair $F'$, $s'$ exists, then return $\bot$ and terminate.
\item\label{alg:match:loop}
$F := F'$; $s := s'$. Repeat from step \ref{alg:match:step-find_as}.
\end{enumerate}
\ifShortVersion
\vspace*{-.1in}
\fi
\end{small}}}
\caption{\texttt{FindMatch} algorithm}
\label{alg:FindMatch}
\end{figure}
To illustrate the  algorithm, let us trace its key parts with $\mathcal{S}_1$ from  Example \ref{ex:date}. Clearly, $\Pi_{AD}(I,\mathcal{F} \setminus \mathcal{D}, \aleph^\times \} \supseteq \{ \hif occurs(d,\mathtt{I}),$ $holds(m,\mathtt{I}), step(\mathtt{I}).$ $forced(m).$ $occurs(d,0).$\}.
Step \ref{alg:match:step-find_ix} infers the initial truth of fluents indirectly from the 
$\mathcal{S}_1$, resulting in an answer set containing $\{ \neg holds(m,$ $0),$ $ forced(m) \}$, i.e., John cannot be initially married. Hence, $I'=I \cup \{ \neg  m \}$. Step \ref{alg:match:step-find_as} checks condition (\ref{crit:1}).
It results in a unique answer set $A \supseteq \{ holds(m,0),$ $\neg holds(ab,0),$ $occurs(d,$ $0),$ $\neg holds(m,1),$ $\neg holds(ab,1) \}$, indicating that $\tbeg \{ \neg m, \neg ab \},$ $d, \{ \neg m, \neg ab \} \tend$ entails $\pm m$.  
Step \ref{alg:match:step-verify} checks condition (\ref{crit:2}). There is a single answer set $B \supseteq \{ u(m,$ $0),$ $\neg holds(ab,0),$ $u(m,1),$ $\neg holds(ab,1) \}$, and, clearly, $\{ holds(m,0),$ $\neg holds(m,0) \} \cap B=\emptyset$. Hence, (\ref{crit:2}) is satisfied
and the
algorithm returns $A$. The semantic score of $\mathcal{S}_1$ is $||A||=0$.

The behavior of the algorithm is characterized by the following theorem, whose proof can be found in \ref{appendix:theorem:algorithm}.
\begin{theorem}\label{theorem:algorithm}
A source $\mathcal{S}$ is a match for a query $\query$ iff \texttt{FindMatch}($I$,$\aleph$,$\query$)$\not=\bot$. The semantic score of $\mathcal{S}$ is $||$\texttt{FindMatch}($I$,$\aleph$,$\query$)$||$.
\end{theorem}

Given a query $\query$ and a collection $\mathcal{S}_1, \ldots, \mathcal{S}_n $ of sources, the Action-Centered \ac{IR} task of ranking the sources based on how relevant each of them is to the problem of finding an answer to $\query$ can now be reduced to (a) using algorithm \texttt{FindMatch} to calculate
$A_i=\texttt{FindMatch}(I_i,\aleph_i,\query)$, where $I_i$ and $\aleph_i$ are the corresponding components of $\mathcal{S}_i$; (b) calculating each semantic score $||A_i||$; and (c) sorting the sources according to their semantic scores.

\subsection{Empirical Evaluation}
While an exhaustive experimental evaluation is beyond the scope of this paper, we include results from a preliminary evaluation we conducted in order to assess the overall viability of our approach. The evaluation is based on a prototypical implementation of \texttt{FindMatch}, which can be downloaded from \url{http://www.mbal.tk/ACIR/}.

It follows immediately from Section \ref{sec:algorithm} that
the execution time of \texttt{FindMatch} only depends on the
source under consideration, which means that the search over
a set of sources can be trivially parallelized.
Note also that the sorting of the 
sources based on their score is clearly dominated by the 
execution time of \texttt{FindMatch}. Thus, the time required for answering a query over a set of
$n$ sources with $m$ identical computing resources is $t\frac{n}{m}$, where $t$ is the average time for processing one source. As a result, in the rest
of this discussion we focus on the execution of \texttt{FindMatch} on individual sources. We organize our evaluation along three dimensions.

\textbf{Sensitivity to problem features.}       We evaluated the sensitivity of \texttt{FindMatch} to variations
        in the problem's features by measuring its performance
        over $100$ problem instances from the \texttt{ins-3-0} set of the Shuttle's Reaction Control System benchmark \citep{bgn07}.
        These instances are significant for at least a preliminary
        evaluation because they involve actions whose effects have intricate
        ramifications, and involve
        the practically-relevant cyberanalytics task of answering questions about 
        a real-world cyber-physical systems.

        For this part of the evaluation, we focused on stories
        consisting of $5$ steps (an intermediate number of steps
        in the context of the original study),
        and potentially up to $3$ concurrent actions per step, for a total of $15$ actions.
        For each selected instance from the original benchmark,
        we randomly generated a sequence of actions of the desired length. The queries were selected in such a way
        that they would lead to a successful match in approximately $50\%$
        of the instances.

The results of the experiment are illustrated in Figure \ref{fig:exp-1} (see \ref{appendix:charts}). The figure reports the execution time for each instance, with the instances colored differently depending on whether they led to a match or not. The corresponding average times are shown as dashed lines. The execution times for instances that led to a match are substantially faster than those of instances that did not, with an average of $0.85$ seconds vs.  $12.81$ seconds. This is not surprising, given that in the latter case the algorithm needs to explore exhaustively all possible forcings and extensions of the action sequence from the source. 

Other than this distinction, it appears that performance of \texttt{FindMatch} is largely independent from the features of the source. In fact, the standard deviation for the ``match'' instances is $1.10$ and for the ``no-match'' instances it is $3.71$.

\textbf{Sensitivity to the number of actions.}  Another aspect of the algorithm that we wanted to evaluate was its sensitivity to the number of actions in a source and, more specifically, to the number of time steps.
        Approaches to reasoning about actions and change
        based on ASP sometimes suffer from a rather 
        substantial growth of the execution time as the number of 
        steps increases.
        Given that the previous experiment demonstrated the
        overall insensitivity of the algorithm to the features
        of the source, for this part of the evaluation we focused on
        the two instances that yielded, respectively, the fastest
        and lowest execution times in the previous experiment.
        
        We randomly generated sequences of actions involving 
        a progressively increasing number of steps, ranging 
        between $3$ and $10$. With up to $3$ parallel actions per step, this yields sources with a number of actions between $9$ and $30$. The outcome of the experiment is illustrated in Figure \ref{fig:exp-2} (see \ref{appendix:charts}). As one might expect, the figure shows an increase of execution time as the number of steps grows. However, the increase is rather moderate, with a worst-case performance of $53.72$ seconds and an average time, across all instances, of $13.08$ seconds. As before, instances that yielded a match were substantially faster than those that did not. Out of $16$ instances, $7$ were solved in less than $1$ second and $10$ in less than $2$ seconds (in fact, in less than $1.10$ seconds).

\textbf{Sensitivity to non-determinism.}        The final aspect of the performance of \texttt{FindMatch} that
        we evaluated is its scalability in the presence of
        actions with non-deterministic effects. As we discussed earlier, determining if a source is a match for a query may require reasoning
        by cases over the effects of non-deterministic
        actions, which tends to increase the number 
        of iterations of \texttt{FindMatch}.
        For this part of the evaluation, we used the same two  instances used in the previous experiment,
        and created $10$ variants for each, so that in
        each variant, $2$ randomly selected actions from
        the story were redefined to have a non-deterministic
        direct effects (as in the first experiment, the number of steps was $5$). The choice to select $2$ actions was
        a pragmatic one, since a source involving a large amount of
        uncertainty is of limited utility in the context of
        Action-Centered \ac{IR}, since the step of manual evaluation
        by a human would require a substantial effort for
        determining the actual effects of the actions. The results of the experiment are shown in Figure \ref{fig:exp-3} (see \ref{appendix:charts}). As one might expect, the execution time in this experiment was higher than in the previous ones, due to the larger number of options \texttt{FindMatch} needs to consider. The maximum execution time was $257.48$ seconds and the minimum $0.81$, with an average of $71.15$ seconds. Out of $20$ instances, $7$ were solved in less than $1$ second, and $9$ in less than $10$ seconds.

\textbf{Overall considerations.} A comprehensive evaluation is needed before general claims can be made, but we believe our experiments show that the approach is promising. In a complex domain such as the Reaction Control System of the Space Shuttle, our simple implementation was able to solve all instances considered in less than $260$ seconds and frequently in less than $1$ second.


\section{Related Work} \label{sec:related-work}
Most traditional \ac{IR} systems base the relevance of a document on a syntactic measurement of the overlap of terms between query and document \citep{MDRS2008}.  Results using this approach may be improved via the application of query expansion \citep{CR2012}, an approach reformulating the original query to expand the sphere of search, for example by collecting synonyms for terms in the query and searching for documents related to those synonyms.

Several approaches for improving search results have been proposed.  
Recent work \citep{RC2012} aims at rethinking the modeling of documents by representing text as a graph whose nodes are terms linked to one another by such properties as co-occurrence in text or grammatical morphology. In this approach, the weights of connections between terms are learned using graph search algorithms such as PageRank \citep{p1999}.  Another interesting area of related research is Temporal IR, or T-IR.  Work in this field aims to improve the
results of Information Retrieval methods by extracting and leveraging temporal information in both documents and queries.  \cite{C2015} presents an extensive survey of the topic.
%
Approaches involving semantic networks, such as Google's Knowledge Graph, bolster \ac{IR} techniques with world facts and relationships. However, they are not concerned with a deeper analysis of query and documents.  

There are a number of research efforts which, while not directly comparable to the work presented here, demonstrate the numerous ways in which \ac{IR} and complex reasoning tasks are being addressed.  One remarkable line of research is that of the question answering agent architecture by \cite{M2016}.  In response to the Facebook set of pre-requisite toy tasks for intelligent question answering \citep{weston2015}, their architecture features a sophisticated reasoning layer that leverages Inductive Logic Programming, implemented in ASP, to learn the knowledge needed to answer the toy task questions.  The authors demonstrated that their agent either matches or outperforms machine learning approaches on the Facebook dataset.  It is important to note that this technique is aimed at question answering, not \ac{IR}.  However, the research on the question answering agent architecture demonstrates the advantages of leveraging formal reasoning for these kinds of tasks, and provides an encouraging indication for our work as well.  

Another approach based on logic and reasoning is in \citep{luka2007}, where the authors aim to answer vague queries such as ``find a car that costs around \$11,000 with about 15,000 miles'' by leveraging description logics and logic programming to rank potential answers by a defined degree of relatedness. Although the notion of degree of relatedness bears some superficial similarities with our work, it should be noted that, once again, this approach is focused on question answering rather than \ac{IR}. Another major difference is that our work aims at reasoning about sequences of events and the effects of those events, both direct and indirect.

\cite{liu2007} presents a novel benchmark dataset for the evaluation of Machine Learning algorithms for ranking text sources in \ac{IR}.  Citing a growing field of feature-based ranking for \ac{IR}, the authors identify and address the lack of standard benchmarks. Although not directly related to our approach, this work may offer useful leads for the creation of evaluation benchmarks.

Finally, \cite{dong2014} propose an approach for the creation of knowledge bases about actions and their effects. They leverage the process of \emph{knowledge fusion}, in which large-scale knowledge bases are automatically extracted from text and associated with a quality measure.

\section{Conclusions and Future Work}\label{sec:conclusions}
In this paper, we presented an investigation of an \ac{IR} task in which sources containing sequences of events are matched to a query about the state of the world after those events.
While this task is critical to simplifying access to information and reducing information overload, traditional \ac{IR} techniques appear unfit to solve it.
Thus, we began by analyzing the problem from a commonsensical and intuitive perspective, and identified characteristics of the corresponding reasoning tasks. We focused particularly on the ability to carry out the fine-grained reasoning needed for a determination of relevance in the presence of uncertainty. Our investigation led us to developing a novel action language, which we used to give an accurate definition of the Action-Centered \ac{IR} task. Finally, we defined an ASP-based procedure for automating the reasoning task and conducted an empirical evaluation of its scalability.

At this stage of our research, we have focused on the definition and study of the core \ac{IR} task. Future work will address the connection with natural language processing and with available knowledge repositories, the development of an end-to-end system, and the quantitative evaluation of our approach. Additionally, it will be interesting to study particular classes of query-source pairs for which simplified forms of reasoning may be possible. For instance, one can check that sources that can be formalized by a deterministic action description without default fluents may be processed without the need for reasoning by cases and, in fact, yield only two possible semantic scores for any query: $0$ and $\infty$. Identifying additional classes may lead to a better understanding of the problem and to more efficient computations.


\bibliographystyle{acmtrans}
\bibliography{biblio-mb,biblio-addon,biblio-el}

\newpage
\appendix
\section{Answer Set Programming}\label{sec:ASP}
ASP \citep{gl91,mt99} is a knowledge representation language with roots in the research on the semantics of logic programming languages and non-monotonic reasoning. The syntax of the language is defined as follows.

Let $\Sigma$ be a signature containing constant, function and
predicate symbols. Terms and atoms are formed as in first-order logic. A \emph{literal} is an atom $a$ or its negation
$\neg a$. 
A \emph{rule} is a statement of the form:
\begin{equation}\label{eq:rule}
h_1 \lor h_2 \ldots \lor h_k \hif l_1, \ldots, l_m, \lpnot l_{m+1}, \ldots, \lpnot l_n
\end{equation}
where each $h_i$  and $l_i$ is a literal and $\mbox{\emph{not}}$ is called
\emph{default negation} operator. The intuitive meaning of \ref{eq:rule} is given in terms of a rational agent reasoning about
its own beliefs and it is summarized by the statement ``a rational agent that believes $ l_1, \ldots, l_m$ and has no reason to believe
$l_{m+1}, \ldots, l_n$, must believe one of $h_1, \ldots, h_k$.''
If $m=n=0$, symbol $\hif$ is omitted and the rule is a \emph{fact}.
%
Rules of the form
$\bot \hif l_1, \ldots, \lpnot l_n$ are abbreviated $\hif l_1, \ldots, \lpnot l_n$, and called \emph{constraints}, intuitively meaning that $\{ l_1, \ldots, \lpnot l_n \}$ must
not be satisfied. A rule with variables is interpreted as
a shorthand for the set of rules obtained by replacing the variables
with all possible variable-free terms.
%
A \emph{program} is a set of rules over $\Sigma$.

Next, we define the semantics of ASP.
We say that a consistent set $S$ of literals is closed under a rule if $\{ h_1, \ldots,  h_k \} \cap S \not= \emptyset$ whenever $\{ l_1, \ldots, l_m \} \subseteq S$ and $\{l_{m+1}, \ldots, l_n\} \cap S = \emptyset$. Set $S$ is an answer set of a \emph{not}-free program $\Pi$ if $S$ is the minimal set closed under its rules. The reduct, $\Pi^S$, of a program $\Pi$ w.r.t. $S$ is obtained from $\Pi$ by removing every rule containing an expression ``$\lpnot l$'' s.t. $l \in S$ and by removing every other occurrence of $\lpnot l$. Set $S$ is an answer set of $\Pi$ if it is the answer set of $\Pi^S$.

\section{Proofs of Theorems}\label{appendix:proofs}
In this appendix, we provide proofs of the main results of this paper. 

\subsection{Proof of Theorem \ref{theorem:asp-and-models}}\label{proof:theorem-1}
Before we proceed to the proof of Theorem \ref{theorem:asp-and-models}, we need to introduce the following notions.
Let $AD$ be an action description of $\mathcal{AL}_{I\!R}$, $n$ be a positive integer, and
$\Sigma(AD)$ be the signature of $AD$. $\Sigma^n(AD)$ denotes the signature
obtained as follows:
\begin{itemize}
\item
$const(\Sigma^n(AD))= const(\Sigma(AD)) \cup \{ 0, \ldots, n \}$
\item
$pred(\Sigma^n(AD))=\{ holds, u, split, occurs \}$
\end{itemize}
Let
\begin{equation}\label{prg:alpha-d}
\alpha^n(AD)=\tbeg \Sigma^n(AD), \Pi^\alpha(AD) \tend,
\end{equation}
where
\begin{equation}\label{prg:pi-alpha-d}
\Pi^\alpha(AD)=\bigcup_{r \in AD} \alpha(r),
\end{equation}
and $\alpha(r)$ is defined as follows:
\begin{itemize}
\item
$\alpha(e \causes \lambda \lif l_1,\ldots,l_n)$ is
\begin{equation}\label{eq:dyn-law-1}
\chi(\lambda,I+1) \hif occurs(e,I), \chi(l_1,I),\ldots, \chi(l_n,I)\mbox{.}
\end{equation}

if $\lambda$ is a fluent literal.  
If $\lambda$ is of the form $u(f)$, the translation of the law is
\begin{gather}
u(f,I+1) \hif occurs(e,I), \chi(l_1,I),\ldots, \chi(l_n,I), \lpnot split(f,I)\mbox{.}\label{pel-1}\\
\chi(f,I+1)\vee \chi(\neg f,I+1) \hif occurs(e,I), \chi(l_1,I),\ldots, \chi(l_n,I), split(f,I).\label{split}
\end{gather}

\item
$\alpha(l_0 \lif l_1,\ldots,l_n)$ is

\begin{equation}
\chi(l_0,T) \hif \chi(l_1,T), \ldots, \chi(l_n,T)\mbox{.}
\end{equation}

\item
$\alpha(e \imp l_1,\ldots,l_n)$ is
\begin{align}
         \hif &\chi(l_1,T), \ldots, \chi(l_n,T), occurs(e,T)\mbox{.}\notag
\end{align}

\end{itemize}

\noindent Let also 
\begin{equation}\label{prg:Phi-d}
\Phi^n(AD)=\tbeg \Sigma^n(AD), \Pi^\Phi(AD) \tend,
\end{equation}
where
\begin{equation}\label{prg:pi-Phi-d}
\Pi^\Phi(AD)=\Pi^\alpha(AD) \cup \Pi
\end{equation}
and $\Pi$ contains the following rules:
\begin{gather}
\chi(F,I+1)\hif \chi(F,I),\lpnot \chi(\neg F,I+1),\lpnot u(F,I+1).\\
\chi(\neg F,I+1)\hif \chi(\neg F,I),\lpnot \chi(F,I+1),\lpnot u(F,I+1).\\
u(F,I+1)\hif u(F,I),\lpnot \chi(F,I+1),\lpnot \chi(\neg F,I+1).
\end{gather}
$\Pi$ also contains the following rules:
\begin{gather}
\hif \chi(F,I),u(F,I).\\
\hif \chi(\neg F,I),u(F,I).
\end{gather}

When we refer to a single action description, we drop argument
$AD$ from the above expressions.

For the rest of this section, we will focus on ground programs. In
order to keep notation simple, we will use $\alpha^n$
and $\Phi^n$ to denote the ground versions of the programs previously defined.

The following notation will be useful in our further discussion. Given a time
point $t$, a state $\sigma$, and a compound action $a$, let
\begin{equation}\label{def:h-o}
\begin{array}{rcl}
\chi(\sigma,t) & = & \{\chi(l,t) \,\,|\,\, l \in \sigma \cap Lit\}\:\cup \\
               &   & \{u(f,t) \,\,|\,\, u(f)\in\sigma\}\\
occurs(a,t) & = & \{occurs(e,t) \,\,|\,\, e \in a \}
\end{array}
\end{equation}
These sets can be viewed as the representation of $\sigma$ and $a$ in ASP.  Let also
\[
split(q_t,t)=\{split(f,t)\,\,|\,\, f\in q_t\}
\]
which represents a set of fluents to which reasoning by cases should be applied according to a qualifier $q_t$.

For any action description $AD$, state $\sigma_0$, and qualified action sequence $s=\tbeg a_0/q_0,\ldots, a_{n-1}/$\\$q_{n-1}\tend$, let $\Phi^n(\sigma_0,s)$ denote
\begin{equation}\label{prg:Phi-d-sigma-a}
\Phi^n \cup \{occurs(a_i,i) \,\,|\,\, a_i \mbox{ is in } s\} \cup \{split(q_i,i) \,\,|\,\, q_i \mbox{ is in } s\}
\end{equation}
Where possible, we drop the first argument, and denote the program by
$\Phi^n(\sigma_0,s)$.  Also, for convenience, we write $\Phi^1(\sigma_0,a_0,q_0)$ when $n=1$.

An important property of $Cn_{Z}$ that we will use later is:

\begin{lemma}\label{u-f-in-succ-state}
For every fluent $f$, $u(f)\in Cn_{Z}$ \emph{iff} $u(f)\in S$.
\Begproof
The thesis follows trivially from the observation that proper extended literals do not occur in state constraints.
\Endproof
\end{lemma}
The following lemma will be helpful in proving the main result of this
section. It states the correspondence between (single) transitions of the
transition diagram and answer sets of the corresponding ASP program.
%

\begin{lemma}\label{lemma:Bdprop}
Let $AD$ be an action description and $\mathcal{T}(AD)$ be the transition diagram it describes. Then, $\tbeg \sigma_0, a_0, \sigma_1 \tend \in
\mathcal{T}(AD)$ iff $\sigma_1=\{l \,\,|\,\, \chi(l,1) \in A\}\cup\{u(f)\,\,|\,\, u(f,1)\in A\}$ for some qualifier $q_0$ and some answer set
$A$ of $\Phi^1(\sigma_0,a_0,q_0)$. 

%

\Begproof
Let us define
\begin{equation}\label{eq:Bdprop:0}
Y_{\sigma_0,a_0,q_0}=\chi(\sigma_0,0) \cup occurs(a_0,0) \cup split(q_0,0)
\end{equation}
and
\[
\Phi^1(\sigma_0,a_0,q_0) = \Phi^1 \cup Y_{\sigma_0,a_0,q_0}
\]

\noindent\emph{Left-to-right}. Let us construct the qualifier $q_0$ as:
\begin{align}\label{qualifier}
q_0=\{f \,\,|\,\,& e \causes u(f) \lif \Gamma\in AD, \\&e\in a_0,
\Gamma\subseteq\sigma_0, \mbox{ and }\notag\\
&u(f)\not\in \sigma_{1}\}\notag
\end{align}
The set $q_0$ is an ASP representation of a qualifier $q_0$ in a qualified action sequence.

Let us show that, if $\tbeg \sigma_0, a_0, \sigma_1 \tend \in
\mathcal{T}(AD)$, then
\begin{equation}\label{eq:Bdprop:constr}
A=Y_{\sigma_0,a_0,q_0} \cup \chi(\sigma_1,1)
\end{equation}
is an answer set of $\Phi^1(\sigma_0,a_0,q_0)$. Notice that $\tbeg \sigma_0, a_0,
\sigma_1 \tend \in \mathcal{T}(AD)$ implies that $\sigma_1$ is a state.  Herein, we refer to $\Phi^1(\sigma_0,a_0,q_0)$ as $P$.

Let us prove that $A$ is the minimal set of literals closed under the rules of the
reduct $P^A$. Let $\mathbb{N}^{\alpha^1(AD)}$ be the set of rules of $\alpha^1(AD)$ of form (\ref{pel-1}).  $P^A$ contains:
\begin{enumerate}
\item[a)]
set $Y_{\sigma_0,a_0,q_0}$.
\item[b)]
all rules in $\alpha^1(AD)\setminus\mathbb{N}^{\alpha^1(AD)}$.
\item[c)] a rule
\[
u(f,1)\hif occurs(e,0),\chi(l_1,0),\ldots,\chi(l_n,0).
\]
for every fluent $f$ such that $split(f,0)\not\in A$.
\item[d)] a rule 
\[
\chi(l,1) \hif \chi(l,0)
\]
for every fluent literal $l$ such that $\chi(l,1) \in A$ and a rule 
\[
\chi(\neg l,1) \hif \chi(\neg l,0)
\]
for every fluent literal $\neg l$ such that $\chi(\neg l,1) \in A$.
\item[e)] a rule 
\[
u(f,1) \hif u(f,0)
\]
for every fluent $f$ such that $u(f,1) \in A$.
\end{enumerate}

Note that because $A$ is an answer set, $\chi(f,1) \in A \Leftrightarrow \chi(\neg f,1)\not\in A \mbox{ and } u(f,1)\not\in A$.  The conditions for $\chi(\neg f)\in A$ and $u(f)\in A$ can be similarly described. 

\bigskip
\noindent\underline{$A$ is closed under $P^A$}. We will prove it for every rule of
the program.

\begin{enumerate}
    \item Rules of groups (a), (d), and (e): obvious.
    \item Rules of group (b) encoding dynamic laws of the form
$e \causes \lambda \lif l_1,\ldots,l_n$ when $\lambda$ is a fluent literal:
\[
\begin{array}{rlcl}
        \chi(\lambda,1)\hif occurs(e,0), \chi(l_1,0), \ldots, \chi(l_n,0).
\end{array}
\]

If $\{o(e,0),\chi(l_1,0), \ldots, \chi(l_n,0)\} \subseteq A$, then, by
(\ref{eq:Bdprop:constr}), $\{l_1,\ldots,l_n\} \subseteq \sigma_0$ and $e \in
a_0$. Therefore, the preconditions of the dynamic law are satisfied by
$\sigma_0$. Hence (\ref{eq:expanded-successor-state}) implies $\lambda \in \sigma_1$. By
(\ref{eq:Bdprop:constr}), $\chi(\lambda,1) \in A$.
\item Rules of group (b) encoding dynamic laws of the form $e \causes \lambda \lif l_1,\ldots,l_n$ when $\lambda$ is of the form $u(f)$:

\[
\chi(f,1)\vee \chi(\neg f,1) \hif occurs(e,0), \chi(l_1,0),\ldots, \chi(l_n,0), split(f,0).
\]
Let us suppose that $split(f,0)\in A$.  In fact, if that is not the case, then $A$ is trivially closed under the rule. Similarly, assume $\{occurs(e,0),\chi(l_1,0),\ldots,\chi(l_n,0)\} \subseteq A$.  Then, by construction of $Y_{\sigma_0,a_0,q_0}$, $split(f,0)\in split(q_0,0)$.
In turn, by construction of $split(q_0,0)$ and from (\ref{qualifier}) we conclude that $f\in q_0$ and that $u(f)\not\in\sigma_1$.  Because $\sigma_1$ is complete from (\ref{eq:classical-successor-state}), we conclude that either $f$ or $\neg f$ is in $\sigma_1$.  By (\ref{eq:Bdprop:constr}), either  $\chi(f,1)\in A$ or $\chi(\neg f,1)\in A$.

\item Rules of group (b) encoding state constraints of the form
$l_0 \lif l_1,\ldots,l_n$:
\[
\begin{array}{rlcl}
        & \chi(l_0,t)        & \hif  & \chi(l_1,t), \ldots, \chi(l_n,t)\mbox{.}
\end{array}
\]
If $\{\chi(l_1,t), \ldots, \chi(l_n,t)\} \subseteq A$, then, by
(\ref{eq:Bdprop:constr}), $\{l_1,\ldots,l_n\} \subseteq \sigma_t$, i.e. the
preconditions of the state constraint are satisfied by $\sigma_t$. If $t=1$, then
(\ref{eq:classical-successor-state}) implies $l_0 \in \sigma_1$. By (\ref{eq:Bdprop:constr}),
$\chi(l_0,t) \in A$. If $t=0$, since states are closed under the state constraints of $AD$,
we have that $l \in \sigma_0$. Again by (\ref{eq:Bdprop:constr}), $\chi(l_0,t) \in
A$.

\item Rules of group (b) encoding executability conditions of the form
$e \imp $\\$l_1,\ldots,l_n$:
\[
\begin{array}{rlcl}
        &               & \hif  & occurs(e,0),\chi(l_1,0), \ldots, \chi(l_n,0).
\end{array}
\]
Since $\tbeg \sigma_0, a_0, \sigma_1 \tend \in \mathcal{T}(AD)$ by hypothesis,
$\tbeg \sigma_0, a_0 \tend$ does not satisfy the preconditions of any
executability condition. Then, either $e \not\in a_0$ or $l_i \not\in \sigma_0$
for some $i$. By (\ref{eq:Bdprop:constr}), the body of this rule is not
satisfied.

\item Rules of group (c) encoding dynamic laws when $\lambda$ is of the form $u(f)$:
\[
u(f,1) \hif occurs(e,0), \chi(l_1,0),\ldots, \chi(l_n,0)\mbox{.}
\]
If the rule is in $P^A$, then $split(f,0)\not\in A$.  By construction of $Y_{\sigma_0,a_0,q_0}$, $split(f,0)\not\in split(q_0,0)$  By construction of $split(q_0,0)$, $f\not\in q_0$ and from (\ref{qualifier}) it follows that $u(f)\in\sigma_1$.  By (\ref{eq:Bdprop:constr}), $u(f,1)\in A$.
\end{enumerate}

\underline{$A$ is the minimal set closed under the rules of $P^A$}. We will
prove this by assuming that there exists a set $B \subseteq A$ such that $B$ is 
closed under the rules of $P^A$, and by showing that $B=A$.

First of all,
\begin{equation}\label{eq:Bdprop:1}
Y_{\sigma_0,a_0,q_0} \subseteq B,
\end{equation}
since these are facts in $P^A$.

Let 
\begin{equation}\label{eq:Bdprop:2}
\delta = \{l \,\,|\,\, \chi(l,1) \in B\}\mbox{.}
\end{equation}
Since $B \subseteq A$,
\begin{equation}\label{eq:Bdprop:3}
\delta \subseteq \sigma_1
\end{equation}
Let $W$ be the element of $\mathbb{E}(a_0,\sigma_0)$ satisfying (\ref{eq:expanded-successor-state}).  We will show that $\delta = \sigma_1$ by proving that
\begin{equation}\label{eq:Bdprop:3-1}
\delta = CN_Z(W \cup (\sigma_1 \cap \sigma_0))\mbox{.}
\end{equation}

\underline{Dynamic laws}. Let $d$ be a dynamic law of $AD$ of the form
$e \causes \lambda \lif l_1,\ldots,l_n$, such that $e \in a_0$ and $\{l_1,\ldots,l_n\}
\subseteq \sigma_0$. Because of (\ref{eq:Bdprop:1}), $\chi(\{l_1,\ldots,l_n\},0)
\subseteq B$ and $o(e,0) \in B$. If $\lambda$ is a fluent literal, then since $B$ is closed under $\alpha(d)$, $\chi(\lambda,1) \in B$, and $\lambda \in \delta$. Therefore,
$W \subseteq \delta$.  It can be similarly shown if $\lambda$ is a properly extended literal.
 
\underline{Inertia}. $P^A$ contains a (reduced) inertia rule of the form
\begin{equation}\label{eq:Bdprop:4}
\chi(f,1) \hif \chi(f,0)\mbox{.}
\end{equation}
for every fluent $f \in \sigma_1$. Suppose $l \in \sigma_1 \cap \sigma_0$.
Then, $\chi(l,0) \in Y_{\sigma_0,a_0,q_0}$, and, since $B$ is closed under (\ref{eq:Bdprop:4}), $\chi(f,1)
\in B$. Therefore, $\sigma_1 \cap \sigma_0 \subseteq \delta$.  The same argument applies to the other reduced inertia rules.

\underline{State constraints}. Let $r$ be a state constraints of $AD$ of the form
$l_0 \lif l_1,\ldots,l_n$, such that
\begin{equation}\label{eq:Bdprop:5}
\chi(\{l_1,\ldots,l_n\},0) \subseteq B\mbox{.}
\end{equation}
Since $B$ is closed under $\alpha(r)$, $\chi(l_0,1) \in B$,
and $l_0 \in \delta$. Then, $\delta$ is closed under the state constraints of $AD$.

Summing up, (\ref{eq:Bdprop:3-1}) holds. From (\ref{eq:expanded-successor-state}) and
(\ref{eq:Bdprop:3}), we obtain $\sigma_1 = \delta$. Therefore $\chi(\sigma_1,1)
\subseteq B$.

At this point we have shown that $Y_{\sigma_0,a_0,q_0} \cup \chi(\sigma_1,1) \subseteq B \subseteq A$.

\emph{Right-to-left}. Let $A$ be an answer set of $P$ and let $\sigma_1=\{l \,\,|\,\,
\chi(l,1) \in A \}\cup\{u(f)\,\,|\,\, u(f,1)\in A\}$. We have to show that
\begin{equation}\label{eq:Bdprop:10}
\sigma_1=CN_Z(W \cup (\sigma_1 \cap \sigma_0)) \mbox{ for some }  W\in\mathbb{E}(a_0,\sigma_0)
\end{equation}
as well as that $\tbeg \sigma_0, a_0 \tend$ respects all executability conditions and that
$\sigma_1$ is consistent and complete.

\underline{$\sigma_1$ consistent}. Obvious, since $A$ is a (consistent) answer set
by hypothesis.

\underline{$\sigma_1$ complete}. By contradiction, and without loss of generality, let $f$ be a fluent s.t. $f
\not\in \sigma_1$, $\neg f \not\in \sigma_1$, $u(f)\not\in\sigma_1$, and $f \in \sigma_0$ (since
$\sigma_0$ is complete by hypothesis, if $f \not\in \sigma_0$, we can still
select $\neg f$ or $u(f)$). Then, the reduct $P^A$ contains a rule
\begin{equation}\label{eq:Bdprop:11}
\chi(f,1) \hif \chi(f,0)\mbox{.}
\end{equation}
Since $A$ is closed under $P^A$, $\chi(f,1) \in A$ and $f \in \sigma_1$.
Contradiction.

\bigskip\bigskip
\underline{Executability conditions respected}. By contradiction, assume that
law $r$ of form  $e \imp$\\$l_1,\ldots,l_n$ is not respected. Note that
$\chi(\{l_1,\ldots,l_n\},0) \subseteq A$ and $occurs(e,0) \in A$. Therefore, the body of
$\alpha(r)$ is
satisfied by $A$, and $A$ is not a answer set.

\underline{(\ref{eq:Bdprop:10}) holds}. Let us construct $W$ so that:
\begin{itemize}
\item
$W \supseteq E(a_0,\sigma_0) \cap Lit$
\item
for every $u(f) \in E(a_0,\sigma_0)$:
\begin{itemize}
\item
if $f \not\in q_0$, then $u(f) \in W$
\item
otherwise, $f \in W$ if $\chi(f,1) \in A$ and $\neg f \in W$ if $\chi(\neg f,1) \in A$.
\end{itemize}
\end{itemize}
One can check that $W \in \mathbb{E}(a_0,\sigma_0)$.

Next, let us prove that $\sigma_1 \supseteq W$, i.e. that for every $\lambda \in W, \lambda \in \sigma_1$. Suppose $\lambda \in E(a_0,\sigma_0) \cap Lit$. There must exist a dynamic law $d$ of the form (\ref{eq:dynamic-law}) such that $\{ l_1, \ldots, l_n \} \subseteq \sigma_0$ and $e \in a_0$. Since $A$ is closed under (\ref{eq:dyn-law-1}) of $\alpha(d)$, it follows that $\chi(\lambda,1)\in A$. By construction of $\sigma_1$, $\lambda \in \sigma_1$.

Let us now consider the case in which $\lambda \not\in E(a_0,\sigma_0) \cap Lit$. There must be a dynamic law $d$ of the form $e \causes u(f) \lif l_1,\ldots,l_n$ such that $f$ is the fluent that occurs in $\lambda$.  It must be the case that $\{ l_1,\ldots,l_n \} \subseteq \sigma_0$, and $e \in a_0$. Note that either $f \in q_0$ or $f \not\in q_0$.

If $f \not\in q_0$, then by construction of $W$ it must be the case that $\lambda$ is $u(f)$. Let us consider (\ref{pel-1}) from $\alpha(d)$. Because $A$ is closed under it, it follows that $u(f,1) \in A$. By construction of $\sigma_1$, we conclude that $u(f)\ \in \sigma_1$.

Next, consider the case in which $f \in q_0$. If $\lambda$ is $f$, then by construction of $W$, one can conclude that $\chi(f,1) \in A$. It follows, then, that $f \in \sigma_1$. If $\lambda$ is $\neg f$, with similar reasoning we derive that $\neg f \in \sigma_1$. This concludes the proof that $\sigma_1 \supseteq W$.

Additionally, $\sigma_1 \supseteq \sigma_1 \cap \sigma_0$ is trivially true.

Let us prove that $\sigma_1$ is closed under the state constraints of $AD$. Consider a
state constraint $s$, of the form $l_0 \lif l_1,\ldots,l_n$, such that $\{l_1,
\ldots, l_n\} \subseteq \sigma_0$. Since $A$ is closed under $\alpha(s)$, $\chi(l_0,1) \in A$. By construction of $\sigma_1$, $l_0 \in \sigma_1$.

Let us prove that $\sigma_1$ is the minimal set satisfying all conditions.
By contradiction, assume that there exists a set $\delta \subset \sigma_1$
such that $\delta \supseteq W \cup (\sigma_1 \cap \sigma_0)$ and
that $\delta$ is closed under the state constraints of $AD$. We will prove
that this implies that $A$ is not an answer set of $P$.

Let $A'$ be the set obtained by removing from $A$ all literals $\chi(l,1)$ such that
$l \in \sigma_1 \setminus \delta$ and all atoms of form $u(f,1)$ such that $u(f) \in \sigma_1 \setminus \delta$. Since $\delta \subset \sigma_1$, $A' \subset
A$.

Since $\delta \supseteq W \cup (\sigma_1 \cap
\sigma_0)$, for every extended fluent literal $\lambda \in \sigma_1 \setminus \delta$ it must be
true that $\lambda \not\in \sigma_0$ and $\lambda \not\in W$.  From Lemma \ref{u-f-in-succ-state}, we conclude that $\lambda$ must be a fluent literal.
Therefore there must exist (at least) one state constraint
$\lambda \lif l_1,\ldots,l_n$ such that $\{l_1, \ldots, l_n \} \subseteq
\sigma_1$ and $\{l_1, \ldots, l_m \} \not\subseteq \delta$. Hence, $A'$ is
closed under the rules of $P^A$. This proves that $A$ is not an answer set of
$P$. Contradiction.
\Endproof
\end{lemma}

\begin{corollary}\label{corollary:1}
Let $AD$ be an action description and $\mathcal{T}(AD)$ be the transition diagram it
describes. Then, $\tbeg \sigma_0, a_0, \sigma_1, \ldots, a_{n-1}, \sigma_n \tend$ is a path of $\mathcal{T}(AD)$ iff 
there exist qualifiers $q_0, q_1, \ldots, q_{n-1}$ and an answer set $A$ of $\Phi^n(\sigma_0,\tbeg a_0/q_0,$ $a_1/q_1,$ $\ldots,$ $a_{n-1}/q_{n-1} \tend)$ such that, for every $1 \leq i \leq n$, 
$\sigma_i=\{l \,\,|\,\, \chi(l,i) \in A\}\cup\{u(f)\,\,|\,\, u(f,i)\in A\}$.

\Begproof
The thesis can be easily proven by induction from Lemma \ref{lemma:Bdprop}.
\Endproof
\end{corollary}

\noindent\emph{Theorem \ref{theorem:asp-and-models}} \\*
Let $I$ be a consistent set of fluent literals, $F$ be a set of fluents, and $s$ be a qualified action sequence. A path $\pi$ is a model of $[ \gamma(I,F)$, $s ]$ iff there exists an answer set of $\Pi_{AD}(I,F,s)$ that encodes $\pi$.
\Begproof
The proof leverages Corollary \ref{corollary:1} and the Splitting Set Lemma \citep{lt94}. First of all, note that it is possible to split $\Pi_{AD}(I,F,s)$ in such a way that the bottom corresponds to rules $[\mathbf{g_1}]$, $[\mathbf{g_2}]$, $[\mathbf{g_3}]$ (see Section \ref{sec:algorithm}) together with facts encoding $I$ and $F$, as well as rules encoding the state constraints for time step $0$. One can check that the answer sets of the bottom encode the completion $\gamma(I,F)$, and that every element of $\gamma(I,F)$ is a state of $\tau(AD)$.

The thesis follows from the application of Corollary \ref{corollary:1} to each $\sigma_0 \in \gamma(I,F)$, after noticing the correspondence between the top of $\Pi_{AD}(I,F,s)$ and program $\Phi^n(\sigma_0,s)$.
\Endproof

\subsection{Proof of Theorem \ref{theorem:algorithm}}\label{appendix:theorem:algorithm}

\noindent \emph{Theorem \ref{theorem:algorithm}} \\
A source $\mathcal{S}$ is a match for a query $\query$ iff \texttt{FindMatch}($I$,$\aleph$,$\query$)$\not=\bot$. The semantic score of $\mathcal{S}$ is $||$\texttt{FindMatch}($I$,$\aleph$,$\query$)$||$.

\Begproof
We begin by showing that the algorithm terminates. This follows simply from the consideration that, in the worst case, the algorithm proceeds to a systematic enumeration of the subsets of $\mathcal{F}$ and of the extensions of $\aleph$ (refer to steps (\ref{alg:match:step-init}), (\ref{alg:match:select}), and (\ref{alg:match:loop})), which are clearly finite, and terminates when all have been enumerated (step (\ref{alg:match:done})).

Next, we demonstrate that if $\Pi_{AD}(I,\mathcal{F} \setminus \mathcal{D},\aleph^\times)$ has at least one answer set, then step (\ref{alg:match:step-find_ix}) of the algorithm finds $\varepsilon(I,\aleph)$, i.e. that $I'=\varepsilon(I,\aleph)$ . Note that the existence of an answer set is verified at step (\ref{alg:match:step-existence}).

\emph{Left-to-right.} Let $A$ be an answer set of $\Pi_{AD}(I,\mathcal{F} \setminus \mathcal{D},\aleph^\times)$. From Theorem \ref{theorem:asp-and-models}, it follows that $A$ encodes a model $\pi_A$ of $[ \gamma(I,\mathcal{F} \setminus \mathcal{D}), \aleph^\times ]$. By construction of $\gamma(I,\mathcal{F} \setminus \mathcal{D})$, 
\begin{equation}\label{eq:num-1}
\mbox{there exists $I' \in I[\mathcal{F} \setminus \mathcal{D}]$ such that $\pi_A$ is a model of $[ \gamma(I'), \aleph^\times ]$.}
\end{equation}
Note that $l \in I'$ iff $l \in I$ or $I \in \{ l' \,|\, \{ \chi(l',0), forced(l'_{f}) \} \subseteq R$, where $l'_{f}$ is the fluent from which $l'$ is formed. If $l \in I$, then from Proposition \ref{prop-2}, $l \in \varepsilon(I,\aleph)$ and the thesis is proven from the observation that the hypothesis of existence of an answer set guarantees the existence of $\varepsilon(I,\aleph)$. In the other case, it follows that $\chi(l,0) \in R$ and that $forced(l_f) \in R$. From the former and (\ref{eq:num-1}), it follows that $l \in \bigcap_{Y \in I[\mathcal{F} \setminus \mathcal{D}]} \gamma(Y)$. Hence,
\begin{equation}\label{eq:num-2}
\mbox{$l \in \gamma(Y)$ for every $Y \in I[\mathcal{F} \setminus \mathcal{D}]$.}
\end{equation}
By construction of $\Pi_{AD}(I,\mathcal{F} \setminus \mathcal{D},\aleph^\times)$, $forced(l_f) \in R$ iff $l_f \in \mathcal{F} \setminus \mathcal{D}$. By definition of forcing of a fluent, every element of $I[\mathcal{F} \setminus \mathcal{D}]$ contains either $l$ or $\overline{l}$. From Proposition \ref{prop-1}, $\gamma(Y)$ is consistent and includes $Y$. From (\ref{eq:num-2}) and the fact that $l \in \gamma(Y)$, it follows that $l \in Y$. Hence, $l \in \bigcap_{Y \in I[\mathcal{F} \setminus \mathcal{D}]} Y$ and thus $l \in \varepsilon(I,\aleph)$. From the generality of $l$, it follows that $I'=\varepsilon(I,\aleph)$.

\emph{Right-to-left.} The conclusion follows from Definition \ref{def:expansion} and Theorem \ref{theorem:asp-and-models} in a straightforward way.

Next, we prove that the algorithm terminates at step (\ref{alg:match:step-return}) iff $\mathcal{S}$ is a match for $\query$. From Theorem \ref{theorem:asp-and-models}, Corollary \ref{corollary:asp-and-models}, and from our observations about step (\ref{alg:match:step-find_ix}), it follows that for every answer set $A$ found at step (\ref{alg:match:step-find_as}), there exists a model $\pi_A$ of $[ \gamma(\varepsilon(I,\aleph),F),s]$ that encodes $A$ and satisfies condition (\ref{crit:1}) of Definition \ref{def:match}. With similar considerations, one can conclude that for every answer set $B$ of $\Pi_{AD}(X,\emptyset,\tbeg\ \tend)$ there exists a model $\pi_B$ of $[ \gamma(\pi_{\sigma_0} \setminus \varepsilon(I,\aleph),\emptyset)$, $\tbeg \  \tend ]$, where $\pi_{\sigma_0}$ is defined in Definition \ref{def:match}. Using Corollary \ref{corollary:asp-and-models}, one can check that the three tests at step (\ref{alg:match:step-verify}) ensure that condition (\ref{crit:2}) from Definition \ref{def:match} is satisfied by $\pi_A$ and $\pi_B$. Thus, if the algorithm terminates at step (\ref{alg:match:step-return}), then $\mathcal{S}$ is a match for $\query$. 

The right-to-left direction is proven by contradiction. We assume that the algorithm never reaches step (\ref{alg:match:step-return}), and yet $\mathcal{S}$ is a match for $\query$. From Definition \ref{def:match}, it follows that there exist $\pi$ and $\pi'$ satisfying conditions (\ref{crit:1}) and (\ref{crit:2}). From Theorem \ref{theorem:asp-and-models} and earlier considerations, it follows that there exist answer sets $A$ and $B$ satisfying the conditions from step (\ref{alg:match:step-find_as}) of the algorithm. This means that the condition of the \emph{if} statement at step (\ref{alg:match:step-return}) is true, and thus the algorithm must terminate, which yields contradiction. This concludes the proof that a source $\mathcal{S}$ is a match for a query $\query$ iff \texttt{FindMatch}($I$,$\aleph$,$\query$)$\not=\bot$.

Next, we demonstrate that the semantic score of $\mathcal{S}$ is $v=||$\texttt{FindMatch}($I$,$\aleph$,$\query$)$||$. If the algorithm returns $\bot$, then $v=\infty$ by definition, and thus the thesis is proven. Otherwise, according to Definition \ref{def:score}, we need to prove that there exist $F$ and\ $s$ such that $v=\Delta(\gamma(\varepsilon(I,\aleph),F))+\Delta(s)$ and that $v$ is minimal among all possible choices of $F$ and $s$ satisfying conditions (\ref{crit:1}) and (\ref{crit:2}) from Definition \ref{def:match}. By construction of $\Pi_{AD}(I,\mathcal{F} \setminus \mathcal{D},\aleph^\times)$, Definition \ref{def:completion}, and the earlier part of the present theorem, it follows that $\Delta(\gamma(\varepsilon(I,\aleph),F))$ is equal to the number of atoms of $A$ formed by relation $forced$. Similarly, $\Delta(s)$ is equal to the number of atoms of $A$ formed by $split$. Hence, $v=\Delta(\gamma(\varepsilon(I,\aleph),F))+\Delta(s)$. The minimality of $v$ is demonstrated by contradiction. Let us proceed by cases. Suppose that, when the algorithm terminates at step (\ref{alg:match:step-return}), $F=\emptyset$ and $s=\aleph^?$. By Definition \ref{def:completion} and Definition \ref{def:match}, $v$ is minimal, which yields contradiction. Suppose, then, that $F\not=\emptyset$ or $s\not=\aleph^?$. Because the values of the two variables are changed only by step \ref{alg:match:loop}, it follows that they were set at that step from the values of $F'$ and $s'$ determined by step \ref{alg:match:select}. However, the values of those variables are selected so that $|F'| + \Delta(s')$ is minimal (step \ref{alg:match:minimality}). Contradiction.
\Endproof

\newpage
\section{Addendum: Figures}\label{appendix:charts}\label{appendix:images}
\begin{figure}[htbp]
\includegraphics[width=.7\columnwidth]{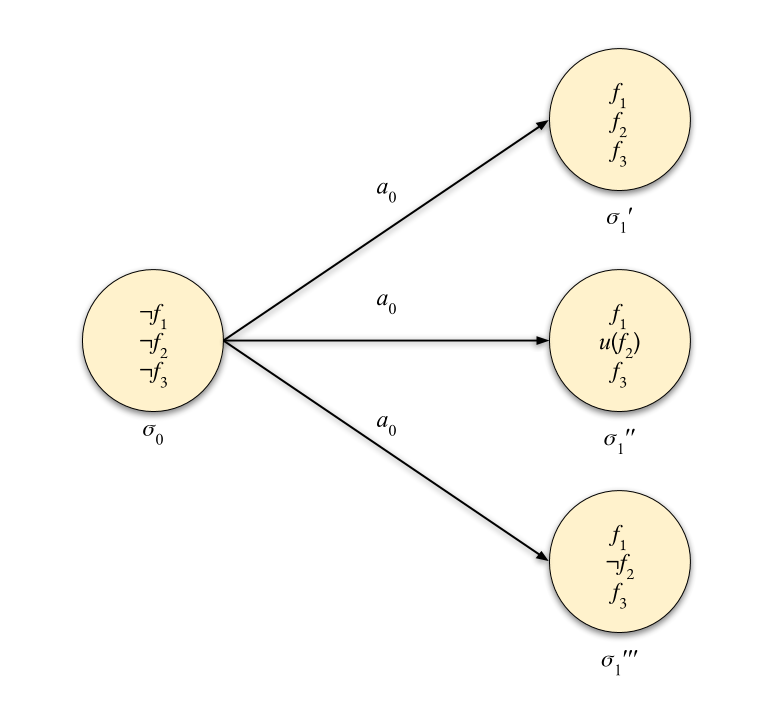}
\caption{Sample transitions}\label{fig:trans-diagram}
\end{figure}
\vspace{2cm}
\begin{figure}[htbp]
\includegraphics[width=\columnwidth]{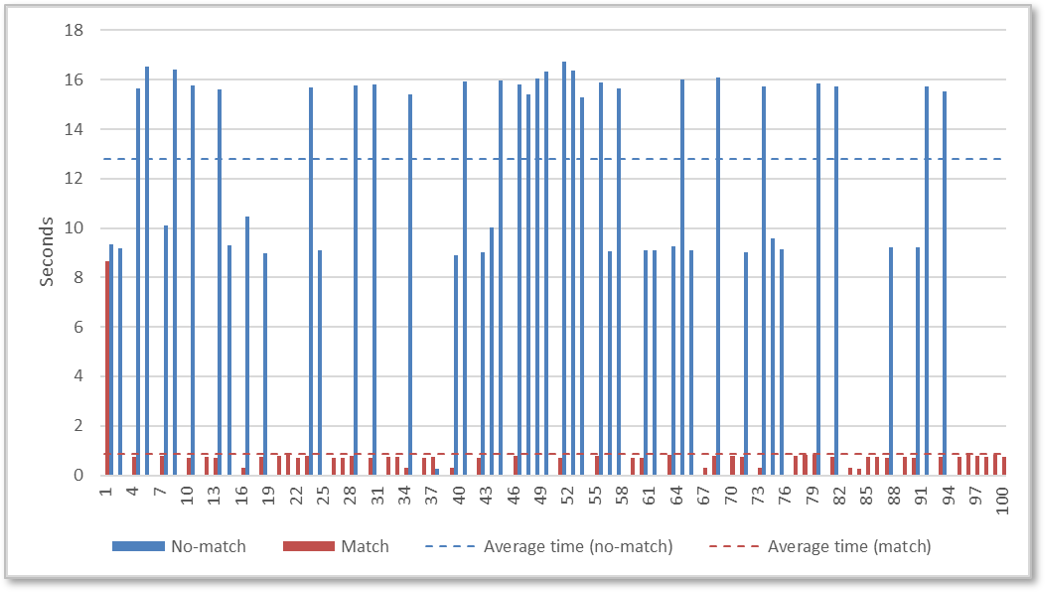}
\caption{Sensitivity to problem features}\label{fig:exp-1}
\end{figure}

\begin{figure}[htbp]
\includegraphics[width=\columnwidth]{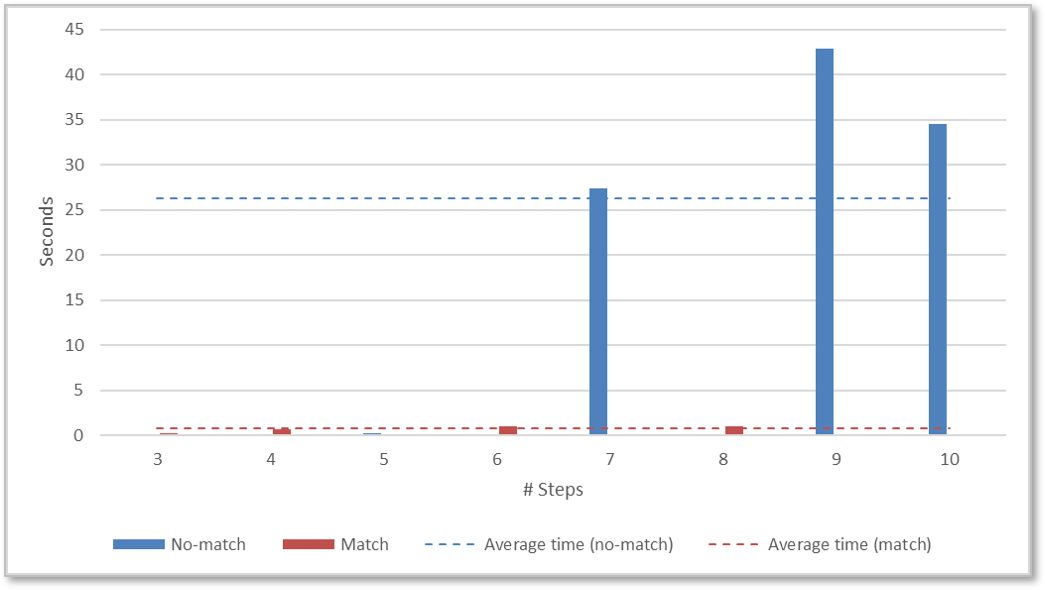}
\includegraphics[width=\columnwidth]{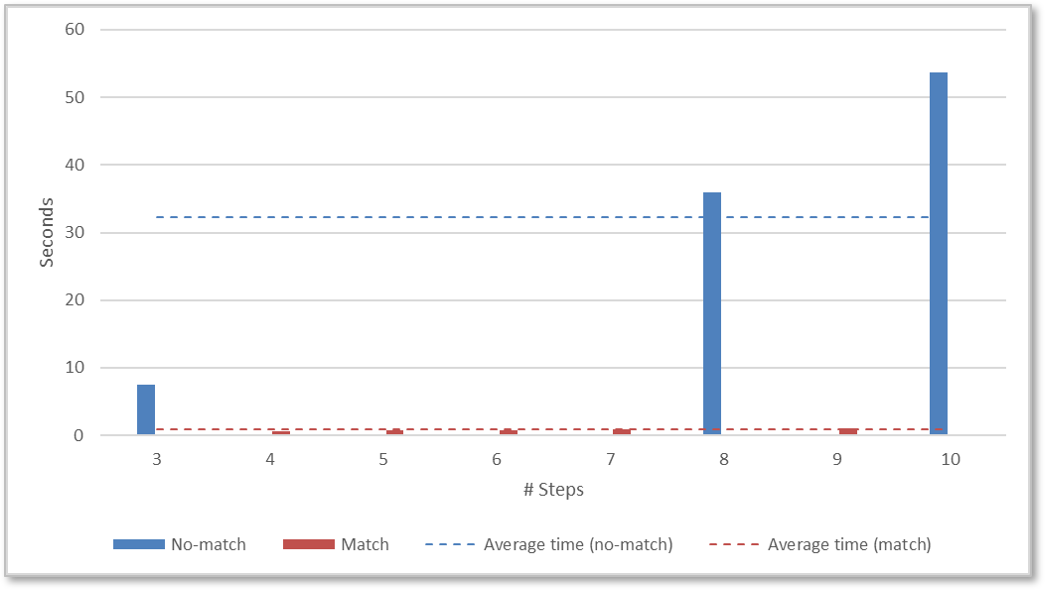}
\caption{Sensitivity to the number of actions, instance set \#1 (top) and instance set \#2 (bottom)}\label{fig:exp-2}
\end{figure}

\begin{figure}[htbp]
\includegraphics[width=\columnwidth]{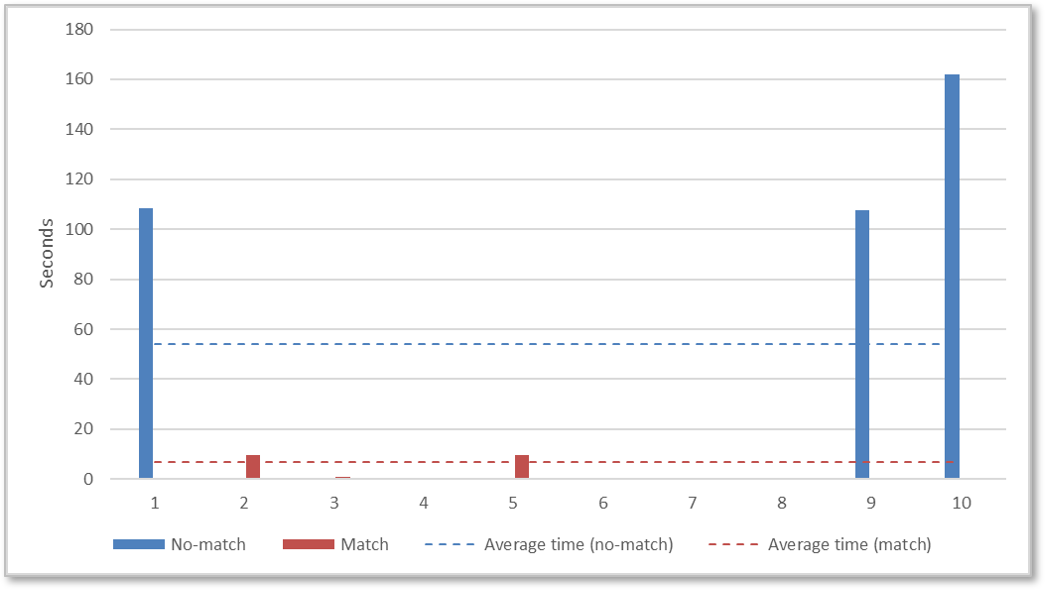} 

\includegraphics[width=\columnwidth]{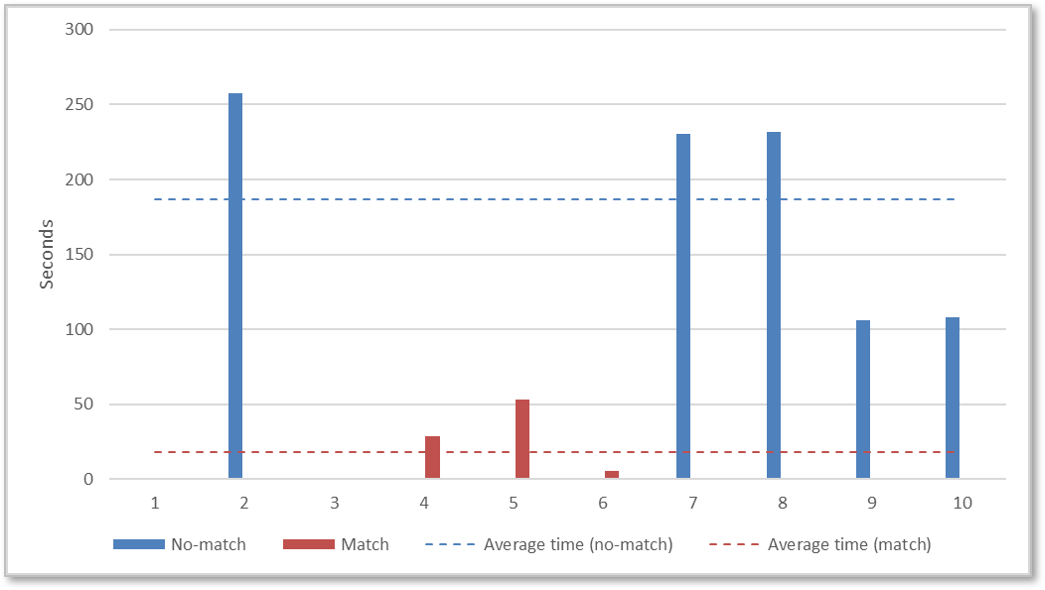}
\caption{Sensitivity to non-determinism, instance set \#1 (top) and instance set \#2 (bottom)}\label{fig:exp-3}
\end{figure}

\end{document}